\documentclass{article} 
\usepackage{iclr2025_conference,times}

\usepackage{hyperref}
\usepackage{url}

\usepackage{amsmath}
\usepackage{amssymb}
\usepackage{mathtools}
\usepackage{amsthm}

\usepackage{mathrsfs}

\usepackage{multirow}

\usepackage[ruled,linesnumbered]{algorithm2e}

\newtheorem{theorem}{Theorem}[section]
\newtheorem{proposition}[theorem]{Proposition}
\newtheorem{lemma}[theorem]{Lemma}
\newtheorem{corollary}[theorem]{Corollary}
\newtheorem{definition}[theorem]{Definition}

\title{Neural Networks on Symmetric Spaces\\ of Noncompact Type}

\author{Xuan Son Nguyen, Shuo Yang, Aymeric Histace  \\
ETIS, UMR 8051, CY Cergy Paris University, ENSEA, CNRS, France\\
\texttt{\{xuan-son.nguyen,shuo.yang,aymeric.histace\}@ensea.fr} \\
}

\newcommand{\olsi}[1]{\,\overline{\!{#1}}}

\iclrfinalcopy 
\begin{document}

\maketitle

\begin{abstract}
Recent works have demonstrated promising performances of neural networks on hyperbolic spaces and 
symmetric positive definite (SPD) manifolds. These spaces belong to a family of Riemannian manifolds 
referred to as symmetric spaces of noncompact type. 
In this paper, we propose a novel approach for developing neural networks on such spaces.  
Our approach relies on 
a unified formulation of the distance from a point to a hyperplane on the considered spaces. 
We show that some existing formulations of the point-to-hyperplane distance can be recovered by our approach under specific settings. 
Furthermore, we derive a closed-form expression for the point-to-hyperplane distance in higher-rank symmetric spaces of noncompact type equipped with $G$-invariant Riemannian metrics. The derived distance then serves as a tool to design
fully-connected (FC) layers and an attention mechanism for neural networks on the considered spaces. 
Our approach is validated on challenging benchmarks for image classification, 
electroencephalogram (EEG) signal classification, image generation, and natural language inference.  
\end{abstract}

\section{Introduction}
\label{sec:intro}

Neural networks in non-Euclidean spaces have become powerful tools for addressing problems in a 
wide range of domains such as natural language processing~\citep{chami2019hyperbolic,NEURIPS2018_dbab2adc}, 
computer vision~\citep{HuangGool17,NguyenHandRecgCVPR19}, 
and medicine~\citep{LiuICML19HGNN}. 
There is a rich existing literature focusing on hyperbolic neural networks (HNNs) due to the ability of hyperbolic spaces to represent 
hierarchical data with high fidelity in low dimensions~\citep{ChamiICML21HoroPCA}. 
Other examples of non-Euclidean spaces which have been commonly encountered 
are SPD manifolds. 
In this paper, we restrict our attention to neural networks with manifold-valued output. 

The concept of hyperplanes has proven useful in the construction of HNNs~\citep{NEURIPS2018_dbab2adc,shimizu2021hyperbolic} and classification algorithms in hyperbolic spaces~\citep{fan2023horospherical}. 
There exist two classes of hyperplanes in hyperbolic spaces, namely, 
Poincar\'e hyperplanes~\citep{NEURIPS2018_dbab2adc,shimizu2021hyperbolic} which are identified as sets of geodesics, 
and horocycles~\citep{fan2023horospherical,helgason1984groups} which are described as 
manifolds orthogonal to families of parallel geodesics. 
Recently, some approaches~\citep{chen2024riemannian,NguyenGyroMatMans23,NguyenICLR24} have successfully developed matrix manifold 
analogs of Poincar\'e hyperplanes. 
However, these approaches either work for SPD manifolds associated with special families of Riemannian metrics~\citep{chen2024riemannian}, 
or require 
rich algebraic structures of the considered spaces, 
which limits their generality. 

In this paper, we present a novel approach for building neural networks on 
symmetric spaces of noncompact type~\citep{helgason1979differential}. 
These include hyperbolic spaces and SPD manifolds and are generally regarded 
as being among the most fundamental and beautiful objects in mathematics~\citep{bridson2011metric,helgason1994geometric}. 
Our contributions are summarized as follows:
\begin{itemize}
\item We propose a novel method to construct the point-to-hyperplane distance in symmetric spaces of 
noncompact type. Compared to~\citet{NEURIPS2018_dbab2adc,NguyenGyroMatMans23,chen2024riemannian} which only concern with hyperbolic spaces~\citep{NEURIPS2018_dbab2adc} or SPD manifolds~\citep{NguyenGyroMatMans23,chen2024riemannian}, our method deals with all those spaces and gives a unified formulation for this distance. 
\item We derive an expression for the point-to-hyperplane distance in a symmetric space of noncompact type equipped with a
$G$-invariant Riemannian metric. 
\item We propose FC layers and an attention mechanism for neural networks on symmetric spaces of noncompact type. 
Within the context of this work, we are the first to develop such building blocks to the best of our knowledge.  
\item We provide experimental results on image classification, EEG signal classification, image generation, and natural language inference showing the efficacy of our approach. 
\end{itemize}

\section{Related Works}
\label{sec:related_works}

\subsection{Hyperbolic Spaces}
\label{subsec:hyperbolic_networks}

HNNs have gained growing attention since the seminal work in~\citet{NEURIPS2018_dbab2adc}, which was inspired by~\citet{LebanonMarginClassifierICML04} and proposed 
hyperbolic analogs of several building blocks of deep neural networks (DNNs). 
Some missing building blocks in~\citet{NEURIPS2018_dbab2adc} (e.g., FC and convolutional layers) 
were then introduced in~\citet{shimizu2021hyperbolic}.
Both the works in~\citet{NEURIPS2018_dbab2adc,shimizu2021hyperbolic} rely primarily on 
the construction of Poincar\'e hyperplanes. 
Another concept of hyperplanes on hyperbolic spaces (horocycles) was studied in~\citet{fan2023horospherical}. 
This approach derives the distance between a point and a horocycle using horospherical projections, which were originally used for dimensionality reduction in hyperbolic spaces~\citep{ChamiICML21HoroPCA}. 
Motivated by the impressive performance of graph neural networks (GNNs)~\citep{velickovic2018graph}, GNNs in hyperbolic spaces were also investigated~\citep{chami2019hyperbolic,gulcehre2018hyperbolic}. 

\subsection{Matrix Manifolds}
\label{subsec:spd_networks}

Most existing works concern with neural networks on SPD and Grassmann manifolds, and special orthogonal groups. 
SPDNet, LieNet, and GrNet were among the first networks designed on those spaces~\citep{HuangGool17,Huang17DLLieGroup,HuangAAAI18}. 
In~\citet{BrooksRieBatNorm19,Ju_2023,kobler2022spdbatchnorm,NguyenFG19,NguyenHandRecgCVPR19,NguyenHandRecgICPR20,Nguyen_2021_ICCV,PanMAttEEG22,WangSymNet21}, 
the authors either introduced Riemannian batch normalization layers or improved Bimap layers~\citep{HuangGool17}.    
The works in~\citet{FedericoGyrocalculusSPD21,NguyenECCV22,NguyenNeurIPS22,NguyenGyroMatMans23,NguyenICLR24} 
leverage rich algebraic structures of SPD and Grassmann manifolds to generalize some basic operations and concepts 
in Euclidean spaces to these manifolds.  
Inspired by~\citet{NguyenGyroMatMans23}, the work in~\citet{chen2024riemannian} generalized multinomial logistic regression (MLR) 
to SPD manifolds under two families of Riemannian metrics. 

\subsection{General Riemannian Manifolds}
\label{subsec:riemannian_networks}

There have also been attempts to develop 
more general frameworks  
for Riemannian manifolds. 
The works in~\citet{ChakrabortyManifoldNet20,zhen2019dilated} advocated the use of weighted Fr\'echet mean to build a number of 
layers (e.g., convolutional and residual layers) for neural networks on Riemannian manifolds. 
In~\citet{katsman2023riemannian}, the authors parameterized vector fields to design 
Riemannian residual neural networks. 
Our work can be connected to this work as one can use our derived distance to parameterize such vector fields. 
Extensions of SPD batch normalization layers~\citep{BrooksRieBatNorm19} on Lie groups were also proposed~\citep{chen2024lie}. 

\section{Mathematical Background}
\label{sec:mathematical_background}

\subsection{Hyperbolic Spaces and SPD Manifolds}
\label{sec:hyperbolic_space_spd_manifolds}

We briefly discuss the geometries of two families of symmetric spaces commonly encountered in machine learning applications.  

\paragraph{Hyperbolic Spaces}
The Poincar\'e model $\mathbb{B}_m$ of $m$-dimensional hyperbolic geometry is defined by the manifold 
$\mathbb{B}_m = \{ x \in \mathbb{R}^m: \| x \| < 1 \}$ equipped with the Riemannian metric 
$\langle u,v \rangle_x = \frac{4}{(1 - \| x \|^2)^2} \langle u,v \rangle$ where $u,v \in \mathbb{R}^m$. 
The Riemannian distance between two points $x,y \in \mathbb{B}_m$ is given by 
$d_{\mathbb{B}}(x,y) = \operatorname{cosh}^{-1}\Big( 1 + 2\frac{\| x-y \|^2}{(1-\| x \|^2)(1-\| y \|^2)} \Big)$. 
A detailed discussion of hyperbolic spaces from a symmetric space perspective is given in Appendix~\ref{sec:supp_hyperbolic_spaces_symspaces}. 

\paragraph{SPD Manifolds}
Here we consider PEM~\citep{chen2024adaptiveLE} (see Appendix~\ref{sec:supp_pem}) which is more general than the well-established 
Log-Euclidean framework~\citep{arsigny:inria-00070423}. Let $\operatorname{Sym}_m$ be the space of $m \times m$ symmetric matrices. 
Under PEM, 
the SPD manifold $\operatorname{Sym}_m^+$ is defined by $\operatorname{Sym}_m^+ = \{ x \in \operatorname{Sym}_m: u^Txu > 0 \text{ for all } u \in \mathbb{R}^m,u \ne \mathbf{0} \}$ equipped with the metric 
$\langle u,v \rangle_x^{\phi} = \langle D_x \phi(u), D_x \phi(v) \rangle$, 
where $\phi: \operatorname{Sym}_m^+ \rightarrow \operatorname{Sym}_m$ is a diffeomorphism, 
$D_x \phi: T_x \operatorname{Sym}_m^+ \rightarrow T_{\phi(x)} \operatorname{Sym}_m$ is the directional derivative of 
map $\phi$ at point $x$, $T_x X$ is the tangent space of $X$ at $x \in X$.  
The Riemannian distance between two points $x,y \in \operatorname{Sym}_m^+$ is given by 
$d_{\mathbb{PEM}}(x,y) = \| \phi(x) - \phi(y) \|$. 
A detailed discussion of SPD manifolds from a symmetric space perspective is given in Appendix~\ref{sec:supp_spd_manifolds_symspaces}. 

Existing point-to-hyperplane distances on Riemannian manifolds are generally built for one of the above families of symmetric spaces,
except for the composite distance~\citep{helgason1984groups,helgason1994geometric}. 
However, the use of the composite distance for our purposes is not straightforward 
since it is a vector-valued distance in higher-rank symmetric spaces (e.g., SPD manifolds). 
In the following, we develop a unified framework to address this limitation of existing works. 

\subsection{Symmetric Spaces of Noncompact Type}
\label{sec:mathematical_background} 

This section briefly recaps important concepts used in the paper. 
We refer the reader to~\citet{ballmann2012,bridson2011metric,helgason1979differential} for further reading. 

Roughly speaking, a symmetric space $X$ is a connected Riemannian manifold which is reflectionally symmetric around any point. 
That is, for any $x \in X$, there exists a local isometry $s_x$ of $X$ such that $s_x(x) = x$ and the differential $D_x s_x = -\operatorname{id}_{T_x X}$. 
Every (simply-connected) symmetric space is a Riemannian product of irreducible symmetric spaces. 
A symmetric space is irreducible, if it cannot be further decomposed into a Riemannian product of symmetric spaces. 
There are two types of (nonflat) irreducible symmetric spaces: compact type and noncompact type. 
Those two types are interchanged by Cartan duality. 
Please refer to Appendix~\ref{sec:supp_symspaces} for further discussion. 
In the following, we restrict our attention to those of noncompact type. 

Formally, let $G$ be a connected noncompact semisimple Lie group with finite center, 
$K$ be a maximal compact subgroup of $G$. 
Then the symmetric space of noncompact type $X$ consists of the left cosets
\begin{equation*}
X := G/K := \{ x = gK | g \in G \}.
\end{equation*}

The action of $G$ on $X = G/K$ is defined as
$g[x] = g[hK] = ghK$ for $x = hK \in X$, $g,h \in G$. 
Let $o$ be the origin $K$ in $X$, 
then the map $\varphi:gK \mapsto g[o]$ 
is a diffeomorphism of $G/K$ onto $X$. 

Let $G=KAN$ be the Iwasawa decomposition~\cite{helgason1979differential,SAWYER2016573} of $G$, 
and let $\mathfrak{g}$ and $\mathfrak{a}$ be the Lie algebras 
of $G$ and $A$, respectively.  
For any linear form $\alpha$ on $\mathfrak{a}$, set $\mathfrak{g}_{\alpha} := \{ v \in \mathfrak{g} | \forall u \in \mathfrak{a}, [u,v] = \alpha(u)v \}$. Let $\mathfrak{a}^{*}$ be the dual space of $\mathfrak{a}$. 
Then the set of restricted roots is defined by $\Sigma := \{ \alpha \in \mathfrak{a}^* \setminus \{ 0 \} | \mathfrak{g}_{\alpha} \ne \{ 0 \} \}$. The kernel of each restricted root is a hyperplane of $\mathfrak{a}$. 
A Weyl chamber in $\mathfrak{a}$ is a connected component of $\mathfrak{a} \setminus \cup_{\alpha \in \Sigma} \operatorname{ker}(\alpha)$. 
We fix a Weyl chamber $\mathfrak{a}^+$ and denote by $\olsi{\mathfrak{a}^+}$ its closure. 

\paragraph{Geometric boundary}
In a symmetric space $X$ of noncompact type, boundary (ideal) points can be regarded as generalizations of the concept of directions in Euclidean spaces. Intuitively, boundary points represent directions along which points in $X$ can move toward infinity~\citep{ChamiICML21HoroPCA}. The set of boundary points $\partial X$ of $X$ is referred to as the (geometric) boundary of $X$. 
For instance, the Poincar\'e disk model (a model of 2-dimensional hyperbolic geometry) is given by 
$\mathbb{D} = \{ (x_1,x_2): x_1^2 + x_2^2 < 1 \}$ (one can think of this set as  the set of all complex numbers with length
less than 1, i.e., $\mathbb{D} = \{ x \in \mathbb{C}: \| x \| < 1 \}$). The boundary $\partial \mathbb{D}$ of $\mathbb{D}$ is the unit circle 
$\partial \mathbb{D} = \{ (x_1,x_2): x_1^2 + x_2^2 = 1 \}$. 
 
Let $d(.,.)$ be the distance induced by the Riemannian metric. 
A geodesic ray in $X$ is a map $\delta: [0, \infty) \rightarrow X$ such that $d(\delta(t),\delta(t')) = |t - t'|,\forall t,t' \ge 0$. 
A geodesic line in $X$ is a map $\delta: \mathbb{R} \rightarrow X$ such that $d(\delta(t), \delta(t')) = |t - t'|, \forall t,t' \in \mathbb{R}$. 
Two geodesic rays $\delta,\delta'$ are said to be asymptotic if $d(\delta(t),\delta'(t))$ is bounded uniformly in $t$. 
This is an equivalence relation on the set of geodesic rays in $X$. 
The set $\partial X$ of boundary points of $X$ 
is the set of equivalence classes of geodesic rays. 
The equivalence class of a geodesic ray $\delta$ is denoted by $\delta(\infty)$. 

\paragraph{Angular metric}

For $x \in X$ and $\xi, \xi' \in \partial X$, there exist unique geodesic rays $\delta$ and $\delta'$ 
which issue from $x$ and lie in the classes $\xi$ and $\xi'$, respectively~\citep{ballmann2012}. 
One can then define $\angle_x(\xi, \xi')$ to be the angle at $x$ between $\delta$ and $\delta'$ (see Appendix~\ref{sec:supp_angles}). 
The angle $\angle(\xi, \xi')$ 
is defined as
\begin{equation*}
\angle(\xi, \xi') = \sup_{x \in X} \angle_x(\xi, \xi'). 
\end{equation*}

The function $(\xi, \xi') \mapsto \angle(\xi, \xi')$ defines the angular metric~\citep{bridson2011metric} on $\partial X$. 



\paragraph{Busemann functions}

Busemann functions (coordinates) can be regarded as generalizations of the concept of coordinates in Euclidean spaces.  
In a Euclidean space, given a point $x$ and a unit vector $w$ (which represents a direction), one has
\begin{equation*}
-\langle x,w \rangle = \lim_{t \rightarrow \infty} (d(x,tw) - d(0,tw)) = \lim_{t \rightarrow \infty} (d(x,tw) - t),
\end{equation*}
where $tw,t > 0$ can be seen as a ray that moves toward infinity in the direction of $w$ as $t \rightarrow \infty$.
Note that the inner product $\langle x,w \rangle$ gives the coordinate of $x$ in the direction of $w$. 
This observation can be used to compute coordinates in $X$.  
Let $\delta:[0, \infty) \rightarrow X$ be a (unit-speed) geodesic ray and $\xi = \delta(\infty) \in \partial X$. 
Then, by replacing $tw$ with geodesic ray $\delta(t)$, one defines the Busemann coordinate of a point $x \in X$ 
in the direction of $\xi$ as
\begin{equation*}
B_{\xi}(x) = \lim_{t \rightarrow \infty} ( d(x, \delta(t)) - t ).
\end{equation*}

The function $B_{\xi}: X \rightarrow \mathbb{R}$ 
is called the Busemann function associated to the geodesic ray $\delta$. 

\paragraph{Horocycles}
Like a Euclidean hyperplane which is orthogonal to a family of parallel lines, a horocycle is orthogonal
to a family of parallel geodesics~\citep{helgason1984groups,helgason1994geometric}. 
Thus, horocycles can be regarded as symmetric space analogs of Euclidean hyperplanes.  
Let $M$ be the centralizer of $A$ in $K$, i.e., 
$M := C_K(A) := \{ k \in K | ka = ak \text{ for all } a \in A \}$. 
The space $\Xi$ of horocycles can be identified~\citep{helgason1994geometric} with
\begin{equation*}
\Xi := G/MN := \{ \eta = gMN | g \in G \}.
\end{equation*} 

\paragraph{Composite distances}

The notion of composite distance is a symmetric space analog of 
the Euclidean inner product~\citep{helgason1984groups,helgason1994geometric}. 
Let $\eta = gMN$ be a horocycle where $g \in G$, and let $g = kan$ where $k \in K$, $a \in A$, and $n \in N$. 
Then $\xi = kM \in \partial X$ is said to be normal to $\eta$, and
$\log(a)$ is the composite distance from the origin $o$ to $\eta$. 
More generally, if $x=gK \in X$, and $\eta = hMN \in \Xi$ where $g,h \in G$, then $H(g^{-1}h)$ is the composite distance from $x$ to $\eta$, 
where the map $H: G \rightarrow \mathfrak{a}$ is determined by $g_1 = k_1 \exp H(g_1) n_1$ with $g_1 \in G$, $k_1 \in K$, and $n_1 \in N$. 

\section{Proposed Approach}
\label{sec:proposed_approach}

We define hyperplanes and propose a general formulation for the point-to-hyperplane distance on the considered spaces 
in Sections~\ref{sec:symspaces_hyperplanes} and~\ref{sec:symspaces_distances_to_hyperplanes}, respectively. 
We then examine the proposed formulation for hyperbolic spaces and SPD manifolds in Section~\ref{subsec:examples}. 
In Section~\ref{subsec:general_formulation_symspace}, our distance is derived for spaces 
equipped with $G$-invariant Riemannian metrics. In Section~\ref{sec:neural_network_symspaces}, we show how to build FC layers 
and an attention mechanism for neural networks on the considered spaces.   

\subsection{Hyperplanes on Symmetric Spaces}
\label{sec:symspaces_hyperplanes}
 
In Euclidean space $\mathbb{R}^m$, a hyperplane $\mathcal{H}^E_{a,b}$ is defined by
\begin{equation*}
\mathcal{H}^E_{a,b} = \{ x \in \mathbb{R}^m: \langle x, a \rangle - b = 0 \},
\end{equation*}
where $a \in \mathbb{R}^m \setminus \{ \mathbf{0} \}$, $b \in \mathbb{R}$, and $\langle .,. \rangle$ is the Euclidean inner product.
The hyperplane $\mathcal{H}^E_{a,b}$ can be reformulated as
\begin{equation*}
\mathcal{H}^E_{a,b} = \{ x \in \mathbb{R}^m: \langle p - x, a \rangle = 0 \},
\end{equation*}
where $p \in \mathbb{R}^m$ and $\langle p, a \rangle = b$.    
 
In order to generalize Euclidean hyperplanes to matrix manifolds, the work in~\citet{NguyenGyroMatMans23} treats parameter $a$ as a point 
on the considered manifold $X$. 
The equation of hyperplane $\mathcal{H}^E_{a,b}$ is then generalized to the matrix manifold setting 
by defining matrix manifold analogs of operations $-$ and $+$ as well as 
that of the Euclidean inner product.  
Here we take a different approach by 
rewriting $\langle p - x, a \rangle$ as a Busemann function. 
Let $\xi$ be the equivalence class of the geodesic ray $\delta(t)=t\frac{a}{\|a\|}$, where  
$\| \cdot \|$ is the Euclidean norm. 
Using the expression of the Busemann function in $\mathbb{R}^m$ 
(see Appendix~\ref{sec:supp_busemann_functions}), 
we have that
\begin{equation*}
\langle p - x, \frac{a}{\|a\|} \rangle = B_{\xi}(-p + x).
\end{equation*}
Assuming that one can define appropriate operations $\ominus$ and $\oplus$ on $X$ 
that are symmetric space analogs of operations $-$ and $+$, respectively. 
This leads us to the following definition. 

\begin{definition}[{\bf Hyperplanes on a Symmetric Space}]\label{def:symspace_hyperplanes}
For $p \in X$ and $\xi \in \partial X$, hyperplanes on $X$ are defined as
\begin{equation*}
\mathcal{H}_{\xi,p} = \{ x \in X: B_{\xi}(\ominus p \oplus x) = 0 \},
\end{equation*}
where $\ominus$ and $\oplus$ are the inverse and binary operations on $X$, respectively.  
\end{definition}

In a symmetric space, a horocycle is a manifold which is orthogonal to families of 
parallel geodesics~\citep{helgason1994geometric}. 
Thus horocycles generalize the idea of Euclidean hyperplanes which are orthogonal to families of parallel lines. 
In our approach, a hyperplane contains a fixed point $p \in X$ and every point $x \in X$ such that 
the segment $\ominus p \oplus x$ is orthogonal to a fixed direction $\xi$. 
Segments of the form $\ominus p \oplus x$ can be regarded as symmetric space analogs of Euclidean lines. 
Therefore, those hyperplanes also generalize the idea of Euclidean hyperplanes in a natural way. 

\begin{figure}[t]
  \begin{center}
    \begin{tabular}{c}      
      \includegraphics[width=0.7\linewidth, trim = 180 280 100 130, clip=true]{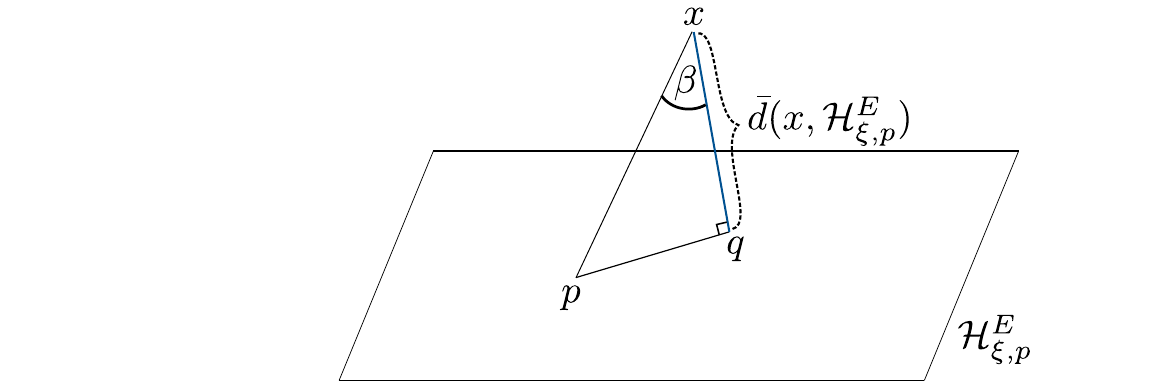} 
    \end{tabular}
  \end{center} 
  \caption{\label{fig:distance_to_hyperplane} The distance between a point $x$ and a hyperplane $\mathcal{H}^E_{\xi,p}$. 
}    
\end{figure}

\subsection{Point-to-hyperplane Distance on Symmetric Spaces}
\label{sec:symspaces_distances_to_hyperplanes}

Let $\mathcal{H}^E_{\xi,p}$ be a hyperplane in $\mathbb{R}^m$. 
Then the distance $\bar{d}(x,\mathcal{H}^E_{\xi,p})$ between a point $x \in \mathbb{R}^m$ 
and $\mathcal{H}^E_{\xi,p}$ can be computed (see Fig.~\ref{fig:distance_to_hyperplane}) as
\begin{equation}\label{eq:point_to_euclidean_hyperplane_distance}
\bar{d}(x,\mathcal{H}^E_{\xi,p}) = d(x,p)\cos(\beta),
\end{equation}
where 
$\beta$ is the angle between the segments $[x,p]$ and $[x,q]$ 
with $q$ being the projection of $x$ on $\mathcal{H}^E_{\xi,p}$. By convention, $\bar{d}(x,\mathcal{H}^E_{\xi,p}) = 0$ for 
any $x \in \mathcal{H}^E_{\xi,p}$. 
Note that Eq.~(\ref{eq:point_to_euclidean_hyperplane_distance}) can be rewritten as
\begin{equation*}
\bar{d}(x,\mathcal{H}^E_{\xi,p}) = d(x,p)\cos \angle_x(\xi',\xi),
\end{equation*}
where $\xi$ and $\xi'$ are the equivalence classes of the geodesic rays $\delta$ and $\delta'$ which issue from 
$x$ and whose images are the segments $[x,q]$ and $[x,p]$, respectively. 
Let $x = \delta(t)$, then
\begin{equation*}
\bar{d}(x,\mathcal{H}^E_{\xi,p}) = d(x,p)\cos \angle_{\delta(t)}(\xi',\xi) = -d(x,p)\lim_{t \rightarrow +\infty} \frac{B_{\xi}(\delta'(t))}{t}. 
\end{equation*}

The last expression~\citep{kapovich2017anosovsubgroupsdynamicalgeometric} is remarkable because it relates the distance $\bar{d}(.,.)$ to a Busemann function. 
Note also that
\begin{equation*}
B_{\xi}(\delta'(t)) = -\langle \delta'(t),a \rangle = -\langle ta',a \rangle,
\end{equation*}
where $\delta(t)=ta,\delta'(t)=ta'$, $a$ is a unit vector, and $a' = \frac{p-x}{\| p-x \|}$. Therefore
\begin{equation*}
\bar{d}(x,\mathcal{H}^E_{\xi,p}) = d(x,p)\langle a',a \rangle = d(x,p)\langle \frac{p-x}{\| p-x \|},a \rangle = d(x,p)\frac{B_{\xi}(-p+x)}{\| -p+x \|}.
\end{equation*}



This motivates the following definition.

\begin{definition}\label{def:distance_to_hyperplane}
Let $\mathcal{H}_{\xi,p}$ be a hyperplane as given in Definition~\ref{def:symspace_hyperplanes}, 
and let $\| \cdot \|_{\mathbb{S}}$ be a norm on $X$. 
Then the (signed) distance $\bar{d}(x,\mathcal{H}_{\xi,p})$ between a point $x \in X$ and $\mathcal{H}_{\xi,p}$ is defined as
\begin{equation*}
\bar{d}(x,\mathcal{H}_{\xi,p}) = d(x,p)\frac{B_{\xi}(\ominus p \oplus x)}{\| \ominus p \oplus x \|_{\mathbb{S}}}.
\end{equation*}
\end{definition}

\subsection{Point-to-hyperplane Distances on Hyperbolic Spaces and SPD Manifolds}
\label{subsec:examples}

We now derive the point-to-hyperplane distance for the symmetric spaces discussed in Section~\ref{sec:hyperbolic_space_spd_manifolds}. 

\paragraph{Hyperbolic Spaces}
The following result is straightforward.
\begin{corollary}\label{corollary:distances_to_hyperplanes_poincare_ball}
Let $\ominus$ and $\oplus$ be the M\"{o}bius subtraction $\ominus_M$ and M\"{o}bius addition $\oplus_M$ in $\mathbb{B}_m$, respectively, 
and let $\| \cdot \|_{\mathbb{S}}$ be the Euclidean norm $\| \cdot \|$ (see Appendix~\ref{sec:mobius_gyrovector_spaces}). 
Let  $p \in \mathbb{B}_m$, $\xi \in \partial \mathbb{B}_m$, 
and let $\mathcal{H}_{\xi,p}$ be a hyperplane as given in Definition~\ref{def:symspace_hyperplanes}. 
Then the distance $\bar{d}(x, \mathcal{H}_{\xi,p})$ between a point $x \in \mathbb{B}_m$ and $\mathcal{H}_{\xi,p}$ is computed by
\begin{equation*}
\bar{d}(x, \mathcal{H}_{\xi,p}) = -\frac{d_{\mathbb{B}}(x,p)}{\| - p \oplus_M x \|} \log \frac{1 - \| - p \oplus_M x \|^2}{\| - p \oplus_M x - \xi \|^2}. 
\end{equation*}
\end{corollary}

\paragraph{SPD Manifolds} 
Proposition~\ref{prop:distances_to_hyperplanes_pullback_metrics} shows that 
the point-to-hyperplane distance studied in~\citet{chen2024riemannian} is a special case of our proposed distance 
(see Appendix~\ref{sec:supp_distances_to_hyperplanes_pullback_metrics} for the proof of Proposition~\ref{prop:distances_to_hyperplanes_pullback_metrics}). 

\begin{proposition}\label{prop:distances_to_hyperplanes_pullback_metrics}
Let $\phi: \operatorname{Sym}_m^+ \rightarrow \operatorname{Sym}_m$ be a diffeomorphism. 
Let $\oplus$ and $\ominus$ be the binary and inverse operations defined by
\begin{equation*}
x \oplus y = \phi^{-1}(\phi(x) + \phi(y)),
\end{equation*}
\begin{equation*}
\ominus x = \phi^{-1}(-\phi(x)),
\end{equation*}
where $x,y \in \operatorname{Sym}_m^+$. 
Let $\| \cdot \|_{\mathbb{S}}$ be the norm induced by the inner product $\langle \cdot \rangle_{\mathbb{S}}$ given as
\begin{equation*}
\langle x,y  \rangle_{\mathbb{S}} = \langle \phi(x),\phi(y) \rangle. 
\end{equation*}
Let $\delta(t) = \phi^{-1}(ta)$ be a geodesic line in $\operatorname{Sym}_m^+$, 
where $a \in \operatorname{Sym}_m$ and $\| a \| = 1$. Let $\xi = \delta(\infty)$, $p \in \operatorname{Sym}_m^+$, 
and let $\mathcal{H}_{\xi,p}$ be a hyperplane as given in Definition~\ref{def:symspace_hyperplanes}. 
Then the distance $\bar{d}(x, \mathcal{H}_{\xi,p})$ between a point $x \in \operatorname{Sym}_m^+$ and $\mathcal{H}_{\xi,p}$ is computed as
\begin{equation*}
\bar{d}(x, \mathcal{H}_{\xi,p}) = \langle a,\phi(p) - \phi(x) \rangle.
\end{equation*}
\end{proposition}
  
A direct consequence of Proposition~\ref{prop:distances_to_hyperplanes_pullback_metrics} is that 
the distance between an SPD matrix and an SPD hypergyroplane~\citep{NguyenGyroMatMans23} is also a special case
of our proposed distance under Log-Euclidean and Log-Cholesky frameworks 
(e.g., the map $\phi$ is the matrix logarithm in the case of Log-Euclidean framework). 

\subsection{Point-to-hyperplane Distance Associated with a $G$-invariant Metric}
\label{subsec:general_formulation_symspace}

In the preceding section, closed-form expressions of the point-to-hyperplane distance 
are computed for hyperbolic spaces and SPD manifolds under PEM. 
In this section, we shall derive this distance in a higher-rank symmetric space $X$ of noncompact type 
equipped with a $G$-invariant Riemannian metric. 
This requires us (1) to define the binary operation $\oplus$ and inverse operation $\ominus$ on $X$; 
(2) to define the norm $\| \cdot \|_{\mathbb{S}}$ on $X$; and (3) to compute the Busemann function. 

Let $x = gK, y = hK \in X$, where $g,h \in G$.

\begin{definition}[{\bf Binary Operation}]\label{def:loop_binary_opt}
The binary operation $\oplus$ is defined as
\begin{equation*}
x \oplus y = ghK.
\end{equation*}
\end{definition}


\begin{definition}[{\bf Inverse Operation}]\label{def:loop_inverse_opt}
The inverse operation $\ominus$ is defined as
\begin{equation*}
\ominus x = g^{-1}K.
\end{equation*}
\end{definition}


The motivation for the above definitions is that the space $G/K$ with the operation $\oplus$ admits 
a group structure (the identity element is $K$ and the inverse of any element is given by the inverse operation). 
In order to compute the norm $\| \cdot \|_{\mathbb{S}}$, we shall define an inner product $\langle \cdot,\cdot \rangle_{\mathbb{S}}$  
whose construction is based on the following natural view points:
\begin{itemize}
\item The inner product $\langle \cdot,\cdot \rangle_{\mathbb{S}}$ should agree with the Riemannian distance. 
\item The inner product $\langle \cdot,\cdot \rangle_{\mathbb{S}}$ should be invariant under the action of $K$. 
This property holds for the ones proposed in~\citet{helgason1994geometric,NguyenGyroMatMans23}. 
\end{itemize}

We thus consider the following inner product. 


\begin{definition}[{\bf The Inner Product on Symmetric Spaces}]\label{def:inner_product_symspace}
Let $x = gK, y = hK \in X$, $g,h \in G$.  
Then the inner product $\langle \cdot,\cdot \rangle_{\mathbb{S}}$ on $X$ is defined as
\begin{equation*}
\langle x,y \rangle_{\mathbb{S}} = \langle \mu(g),\mu(h) \rangle,
\end{equation*}
where the map (Cartan projection) $\mu: G \rightarrow \olsi{\mathfrak{a}^+}$ is determined by $g = k \exp(\mu(g)) k'$ 
with $g \in G$ and $k,k' \in K$ (this follows from the Cartan decomposition~\cite{helgason1979differential} of $G$ where $\mu(\cdot)$ 
 is a continuous, proper, surjective map to the closed Weyl chamber $\olsi{\mathfrak{a}^+}$). 
\end{definition}

Proposition~\ref{lem:main_inner_product_prop} states that the aforementioned properties hold for the considered inner product 
(see Appendix~\ref{sec:supp_main_inner_product_prop} for the proof of Proposition~\ref{lem:main_inner_product_prop}).  
\begin{proposition}\label{lem:main_inner_product_prop}
Let $x = gK, y = hK \in X$, $g,h \in G$, and let $\langle \cdot,\cdot \rangle_{\mathbb{S}}$ be 
the inner product as given in Definitions~\ref{def:inner_product_symspace}. Then

(i) We have that:
\begin{equation*}
\| \ominus x \oplus y \|_{\mathbb{S}} = d(x,y),
\end{equation*} 
where the norm $\|\cdot\|_{\mathbb{S}}$ is induced by the inner product $\langle \cdot,\cdot \rangle_{\mathbb{S}}$. 

(ii) For any $k \in K$, we have that:
\begin{equation*}
\langle x,y \rangle_{\mathbb{S}} = \langle k[x],k[y] \rangle_{\mathbb{S}}.
\end{equation*}

\end{proposition}



Finally, a closed-form expression of the Busemann function is provided in 
Proposition~\ref{lem:busemann_function_symspaces} (see Appendix~\ref{sec:supp_busemann_function_symspaces} for the proof of Proposition~\ref{lem:busemann_function_symspaces}). 

\begin{proposition}\label{lem:busemann_function_symspaces}
Let $\delta(t) = k\exp(ta)K$ be a geodesic ray, 
where $k \in K$, 
$a \in \mathfrak{a}$, $\|a\|=1$, and let $\xi = \delta(\infty)$. Then
\begin{equation*}
B_{\xi}(x) = \langle a,H(g^{-1}) \rangle,
\end{equation*}
where $x \in X$, and $g \in G$ is given by $k^{-1}[x] = gK$. 
\end{proposition}


As a consequence of Proposition~\ref{lem:busemann_function_symspaces}, Corollary~\ref{corollary:distance_to_symspace_hyperplanes} gives 
the expression of the distance between a point and a hyperplane in a symmetric space (see Appendix~\ref{sec:supp_distance_to_symspace_hyperplanes} for the proof of Corollary~\ref{corollary:distance_to_symspace_hyperplanes}).

\begin{corollary}\label{corollary:distance_to_symspace_hyperplanes}
Let $\delta(t) = k\exp(ta)K$ be a geodesic ray, 
where $k \in K$, $a \in \mathfrak{a}$, $\|a\|=1$, and let $\xi = \delta(\infty)$. 
Let $p = hK \in X$, $h \in G$, and let $\mathcal{H}_{\xi,p}$ be a hyperplane given in Definition~\ref{def:symspace_hyperplanes}. 
Then the distance $\bar{d}(x, \mathcal{H}_{\xi,p})$ between a point $x = gK \in X$, $g \in G$ and $\mathcal{H}_{\xi,p}$ is computed as
\begin{equation}\label{eq:connection_to_composite_distance}
\bar{d}(x, \mathcal{H}_{\xi,p}) = \langle a, H(g^{-1}hk) \rangle.
\end{equation} 
\end{corollary}

The connection of the distance in Eq.~(\ref{eq:connection_to_composite_distance}) with existing works is discussed 
in Appendix~\ref{sec:supp_connection_random_features}.  

\section{Neural Networks on Symmetric Spaces}
\label{sec:neural_network_symspaces}


In this section, we shall develop symmetric space analogs of two important building blocks in DNNs, 
i.e., FC layers and attention mechanism. 
Our starting point is the construction of the point-to-hyperplane distance presented in the preceding section. 

\subsection{FC Layers}
\label{sec:fc_layers_symspaces}

An FC layer can be described by the following linear transformation: 
\begin{equation}\label{eq:fc_euclidean}
y = ax - b, 
\end{equation}
where $a \in \mathbb{R}^{m \times m'}$, $x \in \mathbb{R}^{m'}$, and $y,b \in \mathbb{R}^{m}$. Eq.~(\ref{eq:fc_euclidean}) can be rewritten as a system of equations, each for one dimension, i.e., the $j$-th dimension $y_j,j=1\ldots,m$ of the output $y$ is given as
\begin{equation*}
y_j = \langle x,a_j \rangle - b_j, 
\end{equation*} 
where $a_j \in \mathbb{R}^{m'},b_j \in \mathbb{R}$. Let $\tilde{\xi}_j \in \partial X$ and let $\mathcal{H}_{\tilde{\xi}_j,K}$ 
be the hyperplane that contains the origin (i.e., $K$) and is orthonormal to the $j$-th axis of the output space. Then $y_j$ can be interpreted as the signed distance $\bar{d}(y,\mathcal{H}_{\tilde{\xi}_j,K})$ from the output $y$ to hyperplane $\mathcal{H}_{\tilde{\xi}_j,K}$. We thus have
\begin{equation*}
\bar{d}(y,\mathcal{H}_{\tilde{\xi}_j,K}) = \langle x,a_j \rangle - b_j.
\end{equation*}

From Definition~\ref{def:symspace_hyperplanes}, we can write the expression $\langle x,a_j \rangle - b_j$ as $B_{\xi_j}(\ominus p_j \oplus x)$, where $p_j \in X$ and $\xi_j \in \partial X$. Therefore
\begin{equation}\label{eq:fc_layers_basic_equation}
\bar{d}(y,\mathcal{H}_{\tilde{\xi}_j,K}) = B_{\xi_j}(\ominus p_j \oplus x).
\end{equation}
Since the axes of the output space are orthonormal, it is tempting to construct 
a set of orthonormal boundary points $\{ \tilde{\xi}_j \}_{j=1}^m$ for which the output $y$ is related to the 
input $x$ via Eq.~(\ref{eq:fc_layers_basic_equation}).  
Two boundary points $\tilde{\xi}_l$ and $\tilde{\xi}_j,l,j=1,\ldots,m,l \neq j$ are said to be orthonormal 
if $\angle (\tilde{\xi}_l,\tilde{\xi}_j) = \frac{\pi}{2}$. Such a set of boundary points can be identified from 
Proposition~\ref{prop:orthogonal_boundary_points} (see Appendix~\ref{sec:supp_orthogonal_boundary_points} for its proof). 

\begin{proposition}\label{prop:orthogonal_boundary_points}
Let $\delta(t) = \exp(ta)K$ and $\delta'(t) = \exp(ta')K$ be geodesic rays, 
where $a$ and $a'$ are standard basis vectors in $\mathbb{R}^m$, $a \neq a'$. 
Let $\xi = \delta(\infty)$, $\xi' = \delta'(\infty)$. Then $\xi$ and $\xi'$ are orthonormal. 
\end{proposition}


We now formulate our proposed FC layers (see Appendix~\ref{sec:supp_fc_layers_symspaces} for the proof of Proposition~\ref{prop:fc_layers_symspaces}).
\begin{proposition}\label{prop:fc_layers_symspaces}
Let $\delta_j(t) = k_j\exp(ta_j)K,j=1,\ldots,m$ be geodesic rays, where $k_j \in K$, $a_j \in \mathfrak{a}$, $\| a_j \| = 1$. 
Let $v_j(x) = B_{\xi_j}(\ominus p_j \oplus x),j=1,\ldots,m$,  
where $\xi_j = \delta_j(\infty)$, $p_j \in X$, and $x \in X$ is the input of an FC layer. 
Then the output $y$ of the FC layer can be expressed as
\begin{equation*}
y = n \exp([-v_1(x) \ldots -v_m(x)])K,
\end{equation*}
where $n \in N$. 
\end{proposition}


In our approach, the transformation performed by an FC layer is designed to be a symmetric space analog of 
the linear transformation in Eq.~(\ref{eq:fc_euclidean}) which makes our approach distinct from existing ones~\citep{HuangGool17,HuangAAAI18,ChakrabortyManifoldNet20,wang2021laplacian,Sonoda2022FCRidgele}.   
Please refer to Appendix~\ref{sec:supp_fc_distinction} for a comparison of our approach against those approaches. 

\begin{figure*}[t]
  \begin{center}
    \begin{tabular}{c}
      \includegraphics[width=0.7\linewidth, trim = 0 55 0 100, clip=true]{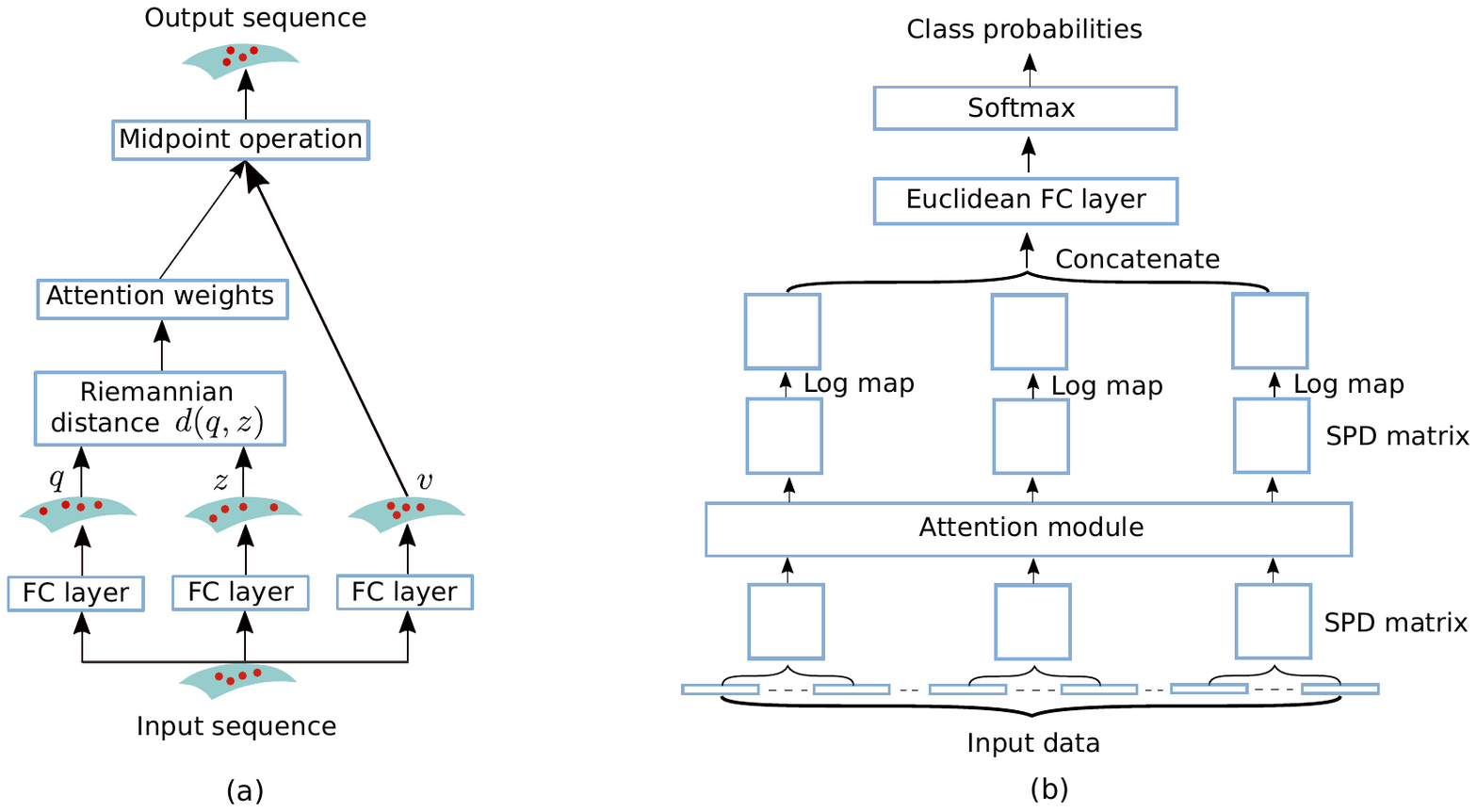} 
    \end{tabular}
  \end{center} 
  \caption{\label{fig:network_architectures} Our proposed attention block (a) and the network architecture for EEG classification (b).}  
\end{figure*} 

\subsection{Attention Mechanism}
\label{sec:attention_symspaces}

We use an approach similar to~\citet{shimizu2021hyperbolic}. 
The scaled dot product attention~\citep{NIPS2017_3f5ee243} is formulated as
\begin{equation}\label{eq:attention}
\operatorname{att}(q,z,v) = \operatorname{softmax}\bigg( \frac{qz^T}{\sqrt{m_z}} \bigg)v,
\end{equation}
where $q, z \in \mathbb{R}^{l \times m_z}$, and $v \in \mathbb{R}^{l \times m_v}$ 
are the queries, keys, and values, respectively, $l$ is the sequence length, 
$m_z$ and $m_v$ are the hidden dimensions of the queries (keys) and values, respectively, 
and function $\operatorname{softmax}(.)$ produces a matrix of the same size as its input matrix 
by applying the softmax function to each row of this matrix. 

The matrix product $qz^T$ corresponds to an attention function that determines the similarities between all query-key pairs. 
The product of function $\operatorname{softmax}(.)$ and $v=[v^T_1;\ldots;v^T_l]$ produces the  
weighted means of values $v_j,j=1,\ldots,l$ and thus can be seen as a midpoint operation.  
In self-attention, the queries, keys, and values are different linear projections of the same input sequence. 
Therefore, Eq.~(\ref{eq:attention}) can be reformulated as
\begin{equation*}
\operatorname{att}(f^q_{lin},f^z_{lin},f^v_{lin},(x_j)^l_{j=1}) = f_{mid}(\{ f^v_{lin}(x_{j}), \pi_{j'j} \}_{j=1}^l)
\end{equation*}
for all $j'=1,\ldots,l$, where $(x_j)^l_{j=1}$ is the input sequence, 
$f^q_{lin}(.)$, $f^z_{lin}(.)$, $f^v_{lin}(.)$ are linear functions that project 
the input points to the queries, keys, and values, respectively, 
$(\pi_{j'j})_{j=1}^l = \operatorname{softmax}\Big( \big(f_{att}(f^q_{lin}(x_{j'}),f^z_{lin}(x_j)) \big)_{j=1}^l \Big)$,  
$f_{att}(.,.)$ is the attention function, 
and $f_{mid}(.)$ is the midpoint operation. 
We use our proposed FC layers (see Fig.~\ref{fig:network_architectures} (a)) to perform linear projections 
in $f^q_{lin}(.)$, $f^z_{lin}(.)$, and $f^v_{lin}(.)$.   
The attention function~\citep{gulcehre2018hyperbolic,shimizu2021hyperbolic} is given as
\begin{equation*}
f_{att}(f^q_{lin}(x_{j'}),f^z_{lin}(x_j)) = -c_1 d(f^q_{lin}(x_{j'}),f^z_{lin}(x_j)) - c_2,
\end{equation*}
where 
$c_1,c_2 \in \mathbb{R}, c_1 > 0$ are learnable parameters. 
We adopt the weighted Fr\'echet mean (wFM) for the midpoint operation. 

\begin{table*}[t]
\caption{\label{tab:exp_distance_expressions} Different formulations of the point-to-hyperplane distance on $\mathbb{B}_m$.  
}
\begin{center}
  \resizebox{1.0\linewidth}{!}{
  \def\arraystretch{1.2}
  \begin{tabular}{| c | c | c |}    
    \hline
    g-distance & h-distance & b-distance \\          
    \hline
    $\operatorname{sinh}^{-1} \Big( \frac{2 | \langle -p \oplus_M x, a \rangle |}{(1 - \| -p \oplus_M x \|^2) \| a \|} \Big)$ & $\frac{1}{a}\big|a \log \frac{1 - \| x \|^2}{\| x - \xi \|^2} - b \big|$ & $-\frac{d_{\mathbb{B}}(x,p)}{\| - p \oplus_M x \|}\log \frac{1 - \| - p \oplus_M x \|^2}{\| - p \oplus_M x - \xi \|^2}$ \\     
    \hline
    $x,p \in \mathbb{B}_m$, $a \in T_p \mathbb{B}_m \setminus \{ \mathbf{0} \}$ & $x \in \mathbb{B}_m$, $a,b \in \mathbb{R},a > 0$, $\xi \in \partial \mathbb{B}_m$ & $x,p \in \mathbb{B}_m$, $\xi \in \partial \mathbb{B}_m$ \\
    \hline
    \citep{NEURIPS2018_dbab2adc} & \citep{fan2023horospherical} & This work \\
    \hline
  \end{tabular}
  } 
\end{center}
\end{table*}

\begin{table}[t]
\caption{\label{tab:exp_resnet_classification} Accuracies (\%) of Hybrid ResNet-18 models for image classification. 
}
\begin{center}
  \resizebox{0.7\linewidth}{!}{
  \def\arraystretch{1.2}
  \begin{tabular}{| l | c | c |}    
    \hline
    Method & CIFAR-10 & CIFAR-100 \\          
    \hline             
    Hybrid Poincar\'e~\citep{guo2022clipped}         & 95.04$\pm$0.13 & 77.19$\pm$0.50 \\
    Poincar\'e ResNet~\citep{vanSpengler2023PResnet} & 94.51$\pm$0.15 & 76.60$\pm$0.32 \\
    Euclidean-Poincar\'e-H~\citep{fan2023horospherical}   & 81.72$\pm$7.84 & 44.35$\pm$2.93 \\ 
    Euclidean-Poincar\'e-G~\citep{NEURIPS2018_dbab2adc}   & 95.14$\pm$0.11 & {\bf 77.78}$\pm$0.09 \\  
    \hline    
    Euclidean-Poincar\'e-B (Ours)                        & {\bf 95.23}$\pm$0.08 & {\bf 77.78}$\pm$0.15 \\ 
    \hline	
  \end{tabular}
  } 
\end{center}
\end{table}

\section{Experiments}
\label{sec:exp}


In this section, we report our experimental evaluation on the image classification 
and EEG signal classification tasks. We refer the reader to Appendix~\ref{sec:experimental_details} for experimental details 
and Appendices~\ref{sec:supp_image_generation} and~\ref{sec:supp_nli} for our experimental evaluation on image generation and natural language inference. 

\subsection{Hyperbolic Spaces}
\label{sec:exp_hnn}  
 
We follow~\citet{BdeirFullyHnnCv24} and design a hybrid architecture\footnote{\url{https://github.com/nguyenxuanson10/symspaces-ic}.} 
which consists of the ResNet-18~\citep{HeResNetCVPR16} and the Poincar\'e MLR~\citep{NEURIPS2018_dbab2adc}. 
The output of the ResNet-18 is mapped to the Poincar\'e ball before it is fed to the Poincar\'e MLR. 
We employ our proposed point-to-hyperplane distance as well as those from~\citet{NEURIPS2018_dbab2adc,fan2023horospherical} (see Tab.~\ref{tab:exp_distance_expressions}) in the Poincar\'e MLR. 
Experiments are conducted on CIFAR-10 and CIFAR-100 datasets~\citep{krizhevsky2009learning}. 
Tab.~\ref{tab:exp_resnet_classification} shows the results of the three resulting networks and those of 
Hybrid Poincar\'e~\citep{guo2022clipped} and Poincar\'e ResNet~\citep{vanSpengler2023PResnet} 
taken from~\citet{BdeirFullyHnnCv24}. 
Hybrid Poincar\'e only differs from Euclidean-Poincar\'e-G in the Poincar\'e MLR which uses
the reparameterization method in~\citet{shimizu2021hyperbolic}. 
Our network gives the best mean accuracies on both datasets. 
In particular, it outperforms all the HNN models from~\citet{BdeirFullyHnnCv24} including the fully hyperbolic model on CIFAR-10 dataset.  
Note that the g-distance is the closest distance from a point to
a Poincar\'e hyperplane, and the b-distance is designed to be a symmetric space analog of the closest distance from a point to a Euclidean hyperplane. However, the h-distance is obtained by horospherical projections which aim to preserve an important property in Principal Component Analysis, i.e., distances between points are invariant to translations along orthogonal directions. 
Therefore, the h-distance does not have the same nature as the g-distance and b-distance. 
This probably explains why Euclidean-Poincar\'e-H is inferior to the other models. 
The inferior performance of the h-distance can also be observed in our experiments for image generation and natural language inference (see Appendices~\ref{sec:supp_image_generation} and~\ref{sec:supp_nli}). 
Furthermore, those experiments demonstrate that: (1) for image generation, the b-distance outperforms the g-distance in 
terms of mean performance in all cases; and (2) for natural language inference, the former 
performs favorably compared to the latter in terms of mean performance in most cases.    
This indicates that our proposed distance has the potential to improve existing HNNs.

\subsection{SPD Manifolds}
\label{sec:exp_spd}

We validate the proposed building blocks (see Section~\ref{sec:neural_network_symspaces}) for SPD neural networks on 
three EEG signal classification datasets: BCIC-IV-2a~\citep{brunner2008BCIComp}, 
MAMEM-SSVEP-II~\citep{Nikolopoulos2021}, and BCI-NER~\citep{PerrinMSBJ12}. 
We test two variants of the network architecture illustrated in Fig.~\ref{fig:network_architectures} (b). 
The FC layers used in the attention block of the first network AttSymSpd-LE 
are built upon Log-Euclidean metrics (see Section~\ref{subsec:examples}), 
while those of the second network AttSymSpd-GI are built upon $G$-invariant metrics 
(see Sections~\ref{subsec:general_formulation_symspace} and~\ref{sec:fc_layers_symspaces}). 

Tab.~\ref{tab:exp_eeg_classification} shows the results of our networks and some state-of-the-art methods. 
Most of these methods are selected~\citep{PanMAttEEG22} based on two criteria: (1) code availability and completeness; 
and (2) solid evaluation (e.g., cross-session) without additional auxiliary procedures. 
As can be observed, AttSymSpd-LE performs the best on all the datasets. 
AttSymSpd-GI is on par with AttSymSpd-LE on BCIC-IV-2a dataset. 
Although AttSymSpd-GI is outperformed by AttSymSpd-LE on MAMEM-SSVEP-II
and BCI-NER datasets, the former enjoys an advantage of having 
much smaller numbers of parameters than the latter. For example, AttSymSpd-GI and AttSymSpd-LE 
use 0.007 MB and 0.034 MB learnable parameters on BCIC-IV-2a dataset, respectively (see also Appendix~\ref{sec:supp_eeg_more_results}). 

\begin{table}[t]
\caption{\label{tab:exp_eeg_classification} Accuracies (\%) of our networks and state-of-the-art methods for EEG signal classification. 
}
\begin{center}
  \resizebox{0.85\linewidth}{!}{
  \def\arraystretch{1.2}
  \begin{tabular}{| l | c | c | c |}    
    \hline
    Method & BCIC-IV-2a & MAMEM-SSVEP-II & BCI-NER \\          
    \hline                 
    EEG-TCNet~\citep{IngolfssonEEGTCNet} & 67.09$\pm$4.6 & 55.45$\pm$7.6 & 77.05$\pm$2.4 \\    
    MBEEGSE~\citep{AltuwaijriMultiBranchCNNEEG22} & 64.58$\pm$6.0 & 56.45$\pm$7.2 & 75.46$\pm$2.3 \\
    MAtt~\citep{PanMAttEEG22} & 74.71$\pm$5.0 & 65.50$\pm$8.2 & 76.01$\pm$2.2 \\
    Graph-CSPNet~\citep{Ju_2023} & 71.95$\pm$13.3 & - & - \\    
    \hline
    AttSymSpd-LE (Ours) & {\bf 78.24} $\pm$ 5.4 & {\bf 70.96} $\pm$ 8.6 & {\bf 78.02} $\pm$ 2.3 \\ 
    AttSymSpd-GI (Ours) & 78.08 $\pm$ 4.8 & 67.24 $\pm$ 7.4 & 75.88 $\pm$ 2.2 \\        
    \hline	
  \end{tabular}
  } 
\end{center}
\end{table}

\section*{Acknowledgements}
We are grateful for the constructive comments and feedback from the anonymous reviewers. 

\bibliography{references}
\bibliographystyle{iclr2025_conference}

\appendix

\section{Implementation Details}
\label{sec:implementation_details}

In this section, we provide details on our implementations of FC layers and the attention module as well as the associated backpropagation procedures.  

\subsection{FC Layers}
\label{sec:details_fc_layers}


{\it Input:} $x \in X$.

{\it Trainable parameters:} $k_j \in K$, $a_j \in \mathfrak{a},\| a_j \|=1$, $p_j \in X,j=1,\ldots,m$, $n \in N$. 

{\it Output:} $y \in X$.

In the case of SPD manifolds, we note that: 
\begin{itemize}
\item $K = O_m$ (the group of $m \times m$ orthogonal matrices). 
\item $A$ is the subgroup of $m \times m$ diagonal matrices with positive diagonal entries.
\item $N$ is the subgroup of $m \times m$ upper-triangular matrices with diagonal entries $1$. 
\end{itemize}

The dimensions of input, output and trainable parameters then can be inferred accordingly ($a_j \in \mathbb{R}^m,j=1,\ldots,m$). 

The computations performed by an FC layer are as follows.

{\it Step 1:} Compute $\ominus p_j \oplus x = h_j^{-1}gK$ where $p_j=h_jK,x=gK,h_j,g \in G,j=1,\ldots,m$. 

{\it Step 2:} Compute $g_j \in G,j=1,\ldots,m$ such that $k_j^{-1}[\ominus p_j \oplus x] = g_jK$. 

{\it Step 3:} Compute $H(g_j^{-1}),j=1,\ldots,m$ from the Iwasawa decomposition of $g_j^{-1}$, i.e., we need to determine the map 
$H: G \rightarrow \mathfrak{a}$ such that $g_j^{-1} = K \exp(H(g_j^{-1})) N$. 

{\it Step 4:} Compute $B_{\xi_j}(\ominus p_j \oplus x),j=1,\ldots,m$ as
\begin{equation*}
B_{\xi_j}(\ominus p_j \oplus x) = \langle a_j, H(g_j^{-1}) \rangle.
\end{equation*}

{\it Step 5:} Compute the output of the FC layer as
\begin{equation*}
y = n \exp \left ( \begin{bmatrix} -B_{\xi_1}(\ominus p_1 \oplus x) & \cdot & \cdot \\ \cdot & -B_{\xi_2}(\ominus p_2 \oplus x) & \cdot \\ \cdot & \cdot & -B_{\xi_m}(\ominus p_m \oplus x) \end{bmatrix} \right ) K
\end{equation*}

\paragraph{Backpropagation}
Below we desribe the backpropagation procedure for FC layers in the case of SPD manifolds. 

{\it Step 1:} Let $p_j = u_js_ju_j^T$ and $x = u_xs_xu_x^T$ be eigen decompositions of $p_j$ and $x$, respectively,   
where $u_j,u_x$ are orthogonal matrices and $s_j,s_x$ are diagonal matrices. Then 
\begin{equation*}
\ominus p_j \oplus x = s_j^{-\frac{1}{2}} u_j^{-1} u_x s_x^{\frac{1}{2}}K.
\end{equation*}  

Eigen decompositions are differentiable operations in the Tensorflow and Pytorch frameworks. 
Please refer to~\citet{Ionescu2015} for a detailed discussion of gradient computations for singular value decompositions and eigen decompositions. 

{\it Step 2:} Compute $g_j = k_j^{-1} s_j^{-\frac{1}{2}} u_j^{-1} u_x s_x^{\frac{1}{2}}$. 

{\it Step 3:} 
Let $g \in G$ and $g = kan$ 
with $k \in K$, $a \in A$, and $n \in N$. Then 
\begin{equation*}
g^Tg = n^Ta^Tk^Tkan = n^Ta^2n = n^Ta(n^Ta)^T. 
\end{equation*} 

Let $g^Tg=cc^T$ be the Cholesky decomposition of $g^Tg$ ($c$ is a lower triangular matrix), 
and let $s$ be the diagonal matrix that contains the main diagonal of $c$. 
Then $n = (cs^{-1})^T$ and $a = s$. The map $H: G \rightarrow \mathfrak{a}$ is then given by $H(g) = \log(a)$. 
We see that the map $H$ can be determined from a Cholesky decomposition which is a differentiable 
operation in the Tensorflow and Pytorch frameworks.  

{\it Step 4:} Backpropagation proceeds as normal with the computation of the inner product $\langle a_j, H(g_j^{-1}) \rangle$. 

{\it Step 5:} The output of the FC layer is computed as
\begin{equation*}
y = n \exp \left ( 2\begin{bmatrix} -B_{\xi_1}(\ominus p_1 \oplus x) & \cdot & \cdot \\ \cdot & -B_{\xi_2}(\ominus p_2 \oplus x) & \cdot \\ \cdot & \cdot & -B_{\xi_m}(\ominus p_m \oplus x) \end{bmatrix} \right ) n^T,
\end{equation*}
which involves only a matrix product. 

\subsection{Attention Module}
\label{sec:details_attention_module}


The pipeline of the attention module is given in Fig.~\ref{fig:network_architectures}(a). 

{\it Input:} A sequence of points $x_j \in X,j=1,\ldots,l$. 

{\it Trainable parameters:} Those include trainable parameters of the three FC layers and $c_1,c_2 \in \mathbb{R}$ (used by the attention function). 

{\it Output:} A sequence of points $y_j \in X,j=1,\ldots,l$. 

In the case of SPD manifolds, the dimensions of input, output and trainable parameters follow from those in FC layers (see above). 

The computations performed by this module are as follows.  

{\it Step 1:} Apply the three FC layers to the sequence of input points $x_j$ and obtain three sequences of points $(q_j)_{j=1}^l$ (queries), 
$(z_j)_{j=1}^l$ (keys), and $(v_j)_{j=1}^l$ (values): 
\begin{equation*}
(q_j)_{j=1}^l = f_{lin}^q((x_j)_{j=1}^l), \hspace{3mm} (z_j)_{j=1}^l = f_{lin}^z((x_j)_{j=1}^l), \hspace{3mm} (v_j)_{j=1}^l = f_{lin}^v((x_j)_{j=1}^l),  
\end{equation*}
where $f_{lin}^q(.)$, $f_{lin}^z(.)$, and $f_{lin}^v(.)$ are the linear transformations performed by the three FC layers. 

{\it Step 2:} Compute the similarities between queries and keys $\bar{\pi}_{j'j},j',j=1,\ldots,l$ from the Riemannian distance function as
\begin{equation*}
\bar{\pi}_{j'j} = -c_1d(q_{j'},z_j) - c_2.
\end{equation*}

{\it Step 3:} Compute the attention weights $\pi_{j'j},j',j=1,\ldots,l$ as
\begin{equation*}
(\pi_{j'j})_{j=1}^l = \operatorname{softmax}\big( (\bar{\pi}_{j'j})_{j=1}^l \big).
\end{equation*}

{\it Step 4:} Perform the midpoint operation to get the sequence of output points $y_{j'},j'=1,\ldots,l$ as
\begin{equation*}
y_{j'} = f_{mid}\big( \{ v_j,\pi_{j'j} \}_{j=1}^l \big),
\end{equation*}
where $\pi_{j'j}$ is the attention weight associated with $v_j$.

\paragraph{Backpropagation}
We desribe the backpropagation procedure for the attention module in the case of SPD manifolds.
Here we only concern with the two blocks 
``Riemannian distance" and ``Midpoint operation". 
For AttSymSpd-GI, the Riemannian distance used in the ``Riemannian distance" block is given by
\begin{equation*}
d(x,y) = \| \log(y^{-\frac{1}{2}}xy^{-\frac{1}{2}}) \|, 
\end{equation*}
where $x,y \in \operatorname{Sym}_m^+$. The term $y^{-\frac{1}{2}}$ is computed as
\begin{equation*}
y^{-\frac{1}{2}} = us^{-\frac{1}{2}}u^T, 
\end{equation*}
where $y = usu^T$ is an eigen decomposition of $y$, $u$ is an orthonormal matrix and $s$ is a diagonal matrix. 
The $\log(.)$ function is also obtained from an eigen decomposition of its argument. 

The ``Midpoint operation" block performs the wFM operation under Log-Euclidean framework. 
Let $\{ x_j,w_j \}_{j=1}^L$ be a set of points $x_j \in \operatorname{Sym}^+_m$ with associated weights 
$w_j \in \mathbb{R}$, where $w_j > 0$ and $\sum_{j=1}^L w_j = 1$. Then the wFM of these points is given by
\begin{equation*}
\operatorname{wFM}(\{ x_j,w_j \}_{j=1}^L) = \exp\bigg(\sum_{j=1}^L w_j \log(x_j)\bigg).
\end{equation*}

The $\exp(.)$ function is computed from an eigen decomposition as $\exp(y) = u\exp(s)u^T$, 
where $u$ is an orthonormal matrix and $s$ is a diagonal matrix. Thus all functions involved in 
the computation of wFM are based on eigen decompositions and so backpropagation proceeds as explained above.

\section{Experimental Details}
\label{sec:experimental_details}

\subsection{Image Classification}
\label{sec:supp_image_classification}

\subsubsection{Datasets}
\label{sec:supp_image_classification_datasets}

\paragraph{CIFAR-10 and CIFAR-100~\citep{krizhevsky2009learning}}
CIFAR-10 and CIFAR-100 datasets contain $60$K $32 \times 32$ colored images from $10$ and $100$ different classes,
respectively. We use the dataset split implemented in PyTorch, which has $50$K training images and $10$K testing images. 

\subsubsection{Experimental Settings}
\label{sec:supp_image_classification_exp_settings}

\paragraph{Network architecture}
Euclidean-Poincar\'e-G, Euclidean-Poincar\'e-B, and Euclidean-Poincar\'e-H have the same architecture which 
consists of the ResNet-18 and the Poincar\'e MLR. Here we only present the Poincar\'e MLR. 
Let $L$ be the number of classes, then MLR computes the probability of each of the output classes as
\begin{align}\label{eq:mlr_reexpression}
\begin{split}
\operatorname{prop}(y=l|x) = \frac{\exp( a_l^Tx  - b_l)}{\sum_{j=1}^L \exp( a_j^Tx  - b_j)} & \propto \exp( a_l^Tx - b_l), 
\end{split}
\end{align}
where $x \in \mathbb{R}^m$ is the input,  
$b_j \in \mathbb{R}$, $a_j \in \mathbb{R}^m, j=1,\ldots,L$ are model parameters. 
One can express~\citep{LebanonMarginClassifierICML04} Eq.~(\ref{eq:mlr_reexpression}) as
\begin{equation}\label{eq:mlr_final_reexpression}
\operatorname{prop}(y=l|x) \propto \exp(\operatorname{sign}( a_l^Tx - b_l) \| a_l \| \bar{d}(x,\mathcal{H}^E_{a_l,b_l})),  
\end{equation}
where $\bar{d}(x,\mathcal{H}^E_{a_l,b_l})$ is the distance between $x$ and hyperplane $\mathcal{H}^E_{a_l,b_l}$ 
(see Section~\ref{sec:symspaces_hyperplanes}).  
In the Poincar\'e MLR~\citep{NEURIPS2018_dbab2adc}, Eq~(\ref{eq:mlr_final_reexpression}) is written as
\begin{equation}\label{eq:poincare_mlr_g}
\operatorname{prop}(y=l|x) \propto \exp\bigg( \frac{2}{(1 - \| p_l \|^2)} \| a_l \| \operatorname{sinh}^{-1} \bigg( \frac{2 | \langle -p_l \oplus_M x, a_l \rangle |}{(1 - \| -p_l \oplus_M x \|^2) \| a_l \|} \bigg) \bigg),
\end{equation}
where $x,p_l \in \mathbb{B}_m$, $a_l \in T_{p_l} \mathbb{B}_m \setminus \{ \mathbf{0} \},l=1,\ldots,L$. 

Euclidean-Poincar\'e-G uses Eq.~(\ref{eq:poincare_mlr_g}) to compute the probability of each of the output classes. 
Euclidean-Poincar\'e-B and Euclidean-Poincar\'e-H are constructed by replacing the point-to-hyperplane distance in Eq.~(\ref{eq:poincare_mlr_g}) with our proposed distance and the one from~\citet{fan2023horospherical}, respectively. 

\paragraph{Hyperparameters}
We follow closely the settings in~\citet{devries2017improvedCNN,BdeirFullyHnnCv24}. 
Random mirroring and cropping are used for training. 
The batch size and number of epochs are set to $128$ and $200$, respectively. 
The learning rate and weight decay are set to $1e-1$ and $5e-4$, respectively. 
The training epochs are set to $60$, $120$, and $160$ for adaptive learning rate scheduling 
where the gamma factor is set to $0.2$. 

\paragraph{Optimization and evaluation}
All models are implemented in Pytorch. We use the library Geoopt~\citep{geoopt2020kochurov} for Riemannian optimization. 
RiemannianSGD is used to train the networks. 
Results are averaged over 5 runs for each model. 
We use a Quadro RTX 8000 GPU for all experiments. 

\begin{table}[t]
\caption{\label{tab:exp_image_classification_more_results} Computation times (seconds) per epoch for image classification experiments (measured on a Quadro RTX 8000 GPU).}
\begin{center}
  \resizebox{0.8\linewidth}{!}{
  \def\arraystretch{1.2}
  \begin{tabular}{| l | c | c | c |}    
    \hline
    Method & Euclidean-Poincar\'e-H & Euclidean-Poincar\'e-G & Euclidean-Poincar\'e-B \\          
    \hline                 
    CIFAR-10  & 52 & 57  & 59 \\ 
    CIFAR-100 & 90 & 114 & 118 \\       
    \hline	
  \end{tabular}
  } 
\end{center}
\end{table}

\subsubsection{More Results}
\label{sec:supp_image_classification_more_results}

The computation times of Euclidean-Poincar\'e-H, Euclidean-Poincar\'e-G and Euclidean-Poincar\'e-B for image classification experiments are
presented in Tab.~\ref{tab:exp_image_classification_more_results}. It can be observed that Euclidean-Poincar\'e-B has high computational costs compared to its competitors, while Euclidean-Poincar\'e-H is the fastest method among the three methods.

\subsection{EEG Signal Classification}
\label{sec:supp_eeg_classification}

\subsubsection{Datasets}
\label{sec:supp_eeg_datasets}

\paragraph{BCIC-IV-2a}
It consists of EEG data captured from $9$ subjects. The cue-based BCI
paradigm consists of $4$ different motor imagery tasks, namely the imagination of movement of the left hand (class $1$), right hand (class $2$), both feet (class $3$), and tongue (class $4$). Two sessions on different days are
recorded for each subject. Each session is comprised of $6$ runs separated by
short breaks. One run consists of $48$ trials ($12$ trials for each of the $4$ possible
classes), yielding a total of $288$ trials per session. 
The signals are recorded with $22$ Ag/AgCl sensors (with inter-electrode distances of $3.5$ cm) 
and sampled at $250$ Hz. 
They are bandpass-filtered between $0.5$ Hz and $100$ Hz. 

\paragraph{MAMEM-SSVEP-II}
It consists of EEG data with $256$ channels captured from $11$ subjects executing a SSVEP-based
experimental protocol. Five different frequencies ($6.66$, $7.50$, $8.57$, $10.00$ and $12.00$
Hz) are used for the visual stimulation, and the EGI $300$ Geodesic EEG
System (GES $300$), using a $256$-channel HydroCel Geodesic Sensor Net (HCGSN)
and a sampling rate of $250$ Hz is used to capture the signals. 

\paragraph{BCI-NER}
It consists of EEG data captured from $26$ subjects.  
The EEG electrode placement follows the extended $10$–$20$ system. 
Five sessions ($60$ trials for the first $4$ sessions and $100$ trials for the last session) are recorded for each subject, 
and the duration of a single EEG trial is $1.25$ seconds.
The signals are recorded with $56$ passive Ag/AgCl sensors (VSM-CTF compatible system) and sampled at $600$ Hz. 
Sixteen subjects released in the early stage of the Kaggle competition\footnote{\url{https://www.kaggle.com/c/inria-bci-challenge}.} are used in our experiments. 


\subsubsection{Experimental Settings}
\label{sec:supp_eeg_exp_settings}

\paragraph{Network architecture}
Inspired by~\citet{HuangGool17}, our network applies a number of convolutional layers to the input data to extract features. The sequence of extracted features is divided into nonoverlapping subsequences, each of them forms an SPD matrix~\citep{HuangGool17}. These procedures create a sequence of SPD matrices, which are fed to the attention block (see Section~\ref{sec:attention_symspaces}). Each output SPD matrix of the attention block is projected 
to the tangent space at the identity matrix via the logarithmic map~\citep{HuangGool17}. The resulting matrices are transformed into vectors, which are then concatenated to create final features for classification. 
The network architecture is illustrated in Fig.~\ref{fig:network_architectures} (b). 

For AttSymSpd-LE, we use the distance derived in Proposition~\ref{prop:distances_to_hyperplanes_pullback_metrics} with $\phi(.) = \log(.)$ 
to build FC layers in the attention module. 
As noted in Section~\ref{subsec:examples}, our definition of hyperplanes and our derived distance 
match the definition of SPD hypergyroplanes and the distance between an SPD matrix and an SPD hypergyroplane, respectively. 
Thus we can use the method in~\citet{NguyenICLR24} for our purposes. 
Let $x \in \operatorname{Sym}^+_{m'}$ be the input of an FC layer, and
let $v_{(l,j)}(x) = \langle \ominus_{le}p_{(l,j)} \oplus_{le} x, a_{(l,j)} \rangle^{le},p_{(l,j)},a_{(l,j)} \in \operatorname{Sym}^+_{m'},l \le j,l,j=1,\ldots,m$ (see Appendix~\ref{sec:le_gyrovector_spaces} for the definitions of operations $\oplus_{le},\ominus_{le}$ and 
the SPD inner product $\langle .,. \rangle^{le}$). 
Then the output $y$ of the FC layer is computed as
\begin{equation*}
y = \exp\big([z_{(l,j)}]_{l,j=1}^{m}\big),
\end{equation*}
where $z_{(l,j)}$ is given by  
\begin{equation*}
z_{(l,j)} = \begin{cases} v_{(l,j)}(x), & \text{if }l=j \\ \frac{1}{\sqrt{2}} v_{(l,j)}(x), & \text{if }l < j \\ \frac{1}{\sqrt{2}} v_{(j,l)}(x), & \text{if }l > j \end{cases}
\end{equation*}

For AttSymSpd-GI, the output $y$ of the FC layer in the attention module is computed as
\begin{equation*}
y = \exp([-v_1(x) \ldots -v_m(x)])K,
\end{equation*}
where $v_j(x) = B_{\xi_j = \delta_j(\infty)}(\ominus p_j \oplus x),p_j \in \operatorname{Sym}^+_{m'},\delta_j(t) = \exp(ta_j)K,j=1,\ldots,m$.
For parameters $p_j = g_jK$, we model them on the space of symmetric matrices, 
and apply the exponential map to obtain SPD matrices~\citep{FedericoGyrocalculusSPD21}. 

The map $H: G \rightarrow \mathfrak{a}$ is computed as follows. Let $g \in G$ and $g = kan$ 
with $k \in K$, $a \in A$, and $n \in N$. Then $g^Tg = n^Ta^Tk^Tkan = n^Ta^2n$, which shows that
$a = \exp(H(g))$ and $n$ can be determined from a LDL decomposition of $g^Tg$. 

To compute the wFM for the midpoint operation, we rely on Log-Euclidean framework.   
Let $\{ x_j,w_j \}_{j=1}^L$ be a set of points $x_j \in \operatorname{Sym}^+_m$ with associated weights 
$w_j \in \mathbb{R}$, where $w_j > 0$ and $\sum_{j=1}^L w_j = 1$. Then the wFM of these points is given by
\begin{equation*}
\operatorname{wFM}(\{ x_j,w_j \}_{j=1}^L) = \exp\bigg(\sum_{j=1}^L w_j \log(x_j)\bigg).
\end{equation*}

\paragraph{Hyperparameters}
To create sequences of SPD matrices for the attention block, the numbers of nonoverlapping subsequences are set to 4, 6, and 4 
on BCIC-IV-2a, MAMEM-SSVEP-II, and BCI-NER datasets, respectively. The number of convolutional layers is set to $2$. 
The numbers of output channels of the first and second convolutional layers are set to $20$ and $15$, respectively. 
The sizes of output SPD matrices of FC layers in the attention block are set to $6 \times 6$, $4 \times 4$, and $4 \times 4$ 
on BCIC-IV-2a, MAMEM-SSVEP-II, and BCI-NER datasets, respectively. 
The numbers of epochs are set to $400$, $100$, and $100$ on BCIC-IV-2a, MAMEM-SSVEP-II, and BCI-NER datasets, respectively. 
The batch sizes are set to $128$, $64$, and $64$ on BCIC-IV-2a, MAMEM-SSVEP-II, and BCI-NER datasets, respectively~\citep{PanMAttEEG22}. 
The learning rate and weight decay are set to $1e-3$ and $1e-1$, respectively. 

\paragraph{Optimization and evaluation}
All models are implemented in Tensorflow. 
Cross-entropy loss and Adam~\citep{KingmaICLR19} are used to train the network.  
Our evaluation protocol is based on~\citet{mane2021fbcnet,PanMAttEEG22,WeiSCCNet19}. 
For BCIC-IV-2a dataset, the session 1 data of a subject is used as the training set 
whose 1/8 is used as the validation set. 
The session 2 data of the same subject is used as the test set. 
For MAMEM-SSVEP-II (BCI-NER) dataset, the first $4$ sessions of a subject are used as the training set 
whose 1/4 is used as the validation set. 
The fifth session of the same subject is used as the test set. 
In all experiments, the models that obtain the lowest losses on the validation sets are used for testing. 
The results on BCIC-IV-2a and MAMEM-SSVEP-II datasets are computed from accuracies obtained over 10 runs for each subject, 
while those on BCI-NER dataset are based on the AUC score.  
Results are averaged over $10$ runs for each model. 
We use a Quadro RTX 8000 GPU for all experiments. 

\begin{table}[t]
\caption{\label{tab:exp_eeg_more_results} Comparison of the numbers of parameters (MB) of AttSymSpd-GI and AttSymSpd-LE.}
\begin{center}
  \resizebox{0.6\linewidth}{!}{
  \def\arraystretch{1.2}
  \begin{tabular}{| l | c | c | c |}    
    \hline
    Dataset & BCIC-IV-2a & MAMEM-SSVEP-II & BCI-NER \\          
    \hline                 
    AttSymSpd-LE & 0.034 & 0.024 & 0.034 \\ 
    AttSymSpd-GI & 0.007 & 0.013 & 0.022 \\       
    \hline	
  \end{tabular}
  } 
\end{center}
\end{table}


\begin{table}[t]
\caption{\label{tab:exp_eeg_block_contributions} Effectiveness of the proposed FC layers and attention module on BCIC-IV-2a dataset.}
\begin{center}
  \resizebox{0.6\linewidth}{!}{
  \def\arraystretch{1.2}
  \begin{tabular}{| l | c | c | c |}    
    \hline
    Method & CovNet & AttSymSpd-GI-Bimap & AttSymSpd-GI \\          
    \hline                 
     & 73.04$\pm$6.34 & 75.82$\pm$5.1 & 78.08$\pm$4.8 \\ 
    \hline	
  \end{tabular}
  } 
\end{center}
\end{table}


\begin{table}[t]
\caption{\label{tab:exp_supp_eeg_classification} Accuracies of our networks and state-of-the-art methods for EEG signal classification. 
}
\begin{center}
  \resizebox{0.9\linewidth}{!}{
  \def\arraystretch{1.2}
  \begin{tabular}{| l | c | c | c |}    
    \hline
    Method & BCIC-IV-2a & MAMEM-SSVEP-II & BCI-NER \\          
    \hline             
    ShallowNet~\citep{Schirrmeister17CNNEEG} & 61.84$\pm$6.39 & 56.93$\pm$6.97 & 71.86$\pm$2.64 \\
    EEGNet~\citep{Lawhern2018} & 57.43$\pm$6.25 & 53.72$\pm$7.23 & 74.28$\pm$2.47 \\
    SCCNet~\citep{WeiSCCNet19} & 71.95$\pm$5.05 & 62.11$\pm$7.70 & 70.93$\pm$2.31 \\
    EEG-TCNet~\citep{IngolfssonEEGTCNet} & 67.09$\pm$4.6 & 55.45$\pm$7.6 & 77.05$\pm$2.4 \\
    TCNet-Fusion~\citep{MUSALLAM2021102826} & 56.52$\pm$3.0 & 45.00$\pm$6.4 & 70.46$\pm$2.9 \\
    FBCNet~\citep{mane2021fbcnet} & 71.45$\pm$4.4 & 53.09$\pm$5.6 & 60.47$\pm$3.0 \\
    MBEEGSE~\citep{AltuwaijriMultiBranchCNNEEG22} & 64.58$\pm$6.0 & 56.45$\pm$7.2 & 75.46$\pm$2.3 \\
    MAtt~\citep{PanMAttEEG22} & 74.71$\pm$5.0 & 65.50$\pm$8.2 & 76.01$\pm$2.2 \\
    Graph-CSPNet~\citep{Ju_2023} & 71.95$\pm$13.3 & - & - \\    
    \hline    
    AttSymSpd-LE (Ours) & {\bf 78.24} $\pm$ 5.4 & {\bf 70.96} $\pm$ 8.6 & {\bf 78.02} $\pm$ 2.3 \\ 
    AttSymSpd-GI (Ours) & 78.08 $\pm$ 4.8 & 67.24 $\pm$ 7.4 & 75.88 $\pm$ 2.2 \\
    \hline	
  \end{tabular}
  } 
\end{center}
\end{table}


\subsubsection{More Results}
\label{sec:supp_eeg_more_results}

Tab.~\ref{tab:exp_eeg_more_results} reports the numbers of learnable parameters of AttSymSpd-GI and AttSymSpd-LE. 
Results clearly show that AttSymSpd-GI uses far fewer parameters than AttSymSpd-LE. 
It is interesting to note that these networks give similar accuracies on BCIC-IV-2a dataset, 
but AttSymSpd-GI has about $5 \times$ fewer parameters than AttSymSpd-LE.  

We also study the effectiveness of the proposed FC layers and attention module for EEG signal classification. 
To this end, we evaluate the performance of AttSymSpd-GI in two cases:
\begin{itemize}
\item The attention module is removed from the network. The resulting network is called CovNet. This allows us to validate the contribution of the attention module. 
\item The FC layers in the attention module are replaced with Bimap layers~\citep{HuangGool17}. Bimap layers are referred to as FC convolution-like layers and arguably the most commonly used analogs of FC layers in SPD neural networks. The resulting network is called AttSymSpd-GI-Bimap. 
\end{itemize}

Tab.~\ref{tab:exp_eeg_block_contributions} reports results of our experiments. It can be observed that both the building blocks are effective. In particular, the use of attention module leads to more than 5\% improvement in mean accuracy, and the network based on our FC layers outperforms the one based on Bimap layers by more than 2\% in terms of mean accuracy.

Finally, Tab.~\ref{tab:exp_supp_eeg_classification} shows all the results from Tab.~\ref{tab:exp_eeg_classification} and 
additional results of some state-of-the-art methods. It can be seen that AttSymSpd-LE outperforms state-of-the-art methods
on all the datasets, while AttSymSpd-GI outperforms them on BCIC-IV-2a and MAMEM-SSVEP-II datasets.

\begin{table}[t]
\caption{\label{tab:network_architecture_image_generation} The network architecture for image generation. 
The PROJ$_{\mathbb{R}^m \rightarrow \mathbb{L}_m}$ layer maps data in $\mathbb{R}^m$ to $\mathbb{L}_m$. 
The H-PROJ$_{\mathbb{L}_m \rightarrow \mathbb{B}_m}$ and H-PROJ$_{\mathbb{B}_m \rightarrow \mathbb{L}_m}$ layers
map data between $\mathbb{L}_m$ and $\mathbb{B}_m$ (see the text). 
The CONV and CONVTR layers are Lorentz analogs of the convolutional and transposed convolutional layers, respectively. 
The BN and RELU layers are Lorentz analogs of the batch normalization and ReLU layers, respectively. 
The FC-MEAN, FC-VAR, and FC layers are Lorentz analogs of the Euclidean FC layer, respectively. 
The SAMPLE layer generates random samples in $\mathbb{B}_m$ from the latent distribution of the network. 
The MLR layer is the Poincar\'e MLR. 
Convolutional layers and transposed convolutional layers have kernel sizes of $3 \times 3$ and of $4 \times 4$, respectively.  
$s$ and $p$ denote stride and zero padding, respectively. 
}
\begin{center}
  \resizebox{0.7\linewidth}{!}{
  \def\arraystretch{1.2}
  \begin{tabular}{| l | r |}    
    \hline
    Layer & CIFAR-10 / CIFAR-100 \\          
    \hline             
    ENCODER: & \\
    $\rightarrow$ PROJ$_{\mathbb{R}^m \rightarrow \mathbb{L}_m}$ & $8 \times 8 \times 3$ \\
    $\rightarrow$ CONV$_{65,s2,p1}$ $\rightarrow$ BN $\rightarrow$ RELU & $4 \times 4 \times 65$ \\
    $\rightarrow$ FLATTEN & 1025 \\
    $\rightarrow$ FC-MEAN$_{129}$ & 129 \\
    $\rightarrow$ FC-VAR$_{129}$ $\rightarrow$ SOFTPLUS & 129 \\
    \hline
    DECODER: & \\
    $\rightarrow$ H-PROJ$_{\mathbb{L}_m \rightarrow \mathbb{B}_m}$ & 128 \\
    $\rightarrow$ SAMPLE ($\mathbb{B}_m$) & 128 \\
    $\rightarrow$ H-PROJ$_{\mathbb{B}_m \rightarrow \mathbb{L}_m}$ & 129 \\
    $\rightarrow$ FC$_{257}$ $\rightarrow$ BN $\rightarrow$ RELU & 257 \\
    $\rightarrow$ RESHAPE & $2 \times 2 \times 65$ \\
    $\rightarrow$ CONVTR$_{33,s2,p1}$ $\rightarrow$ BN $\rightarrow$ RELU & $4 \times 4 \times 33$ \\
    $\rightarrow$ CONVTR$_{17,s2,p1}$ $\rightarrow$ BN $\rightarrow$ RELU & $8 \times 8 \times 17$ \\
    $\rightarrow$ CONV$_{65,s1,p1}$ & $8 \times 8 \times 65$ \\
    $\rightarrow$ H-PROJ$_{\mathbb{L}_m \rightarrow \mathbb{B}_m}$ & $8 \times 8 \times 64$ \\
    $\rightarrow$ MLR ($\mathbb{B}_m$) & $8 \times 8 \times 3$ \\ 
    \hline	
  \end{tabular}
  } 
\end{center}
\end{table}

\section{Image Generation}
\label{sec:supp_image_generation} 

In this section, we perform image generation experiments using CIFAR-10 and CIFAR-100 datasets. 
We design a new hyperbolic variational autoencoder (VAE) from HCNN Lorentz~\citep{BdeirFullyHnnCv24} in which 
we replace the hyperbolic wrapped normal distribution in the Lorentz model with that in the Poincar\'e ball,
and replace the Lorentz MLR with the Poincar\'e MLR. 
We use three different point-to-hyperplane distances in the Poincar\'e MLR as in our image classification experiments.  

\subsection{The Lorentz Model}
\label{sec:supp_lorentz_model}

The Lorentz model $\mathbb{L}_m$ of $m$-dimensional hyperbolic geometry is defined by the manifold 
$\mathbb{L}_m = \{ x = [x_0,\ldots,x_m]^T \in \mathbb{R}^{m+1}, x_0 > 0: -x_0^2 + \sum_{i=1}^m x_i^2 = -1 \}$ 
equipped with the Riemannian metric $\langle .,. \rangle_x = \operatorname{diag}(-1,\ldots,1)$. 
The Riemannian distance between two points $x = [x_0,\ldots,x_m]^T, y = [y_0,\ldots,y_m]^T \in \mathbb{L}_m$ is given by 
$d_{\mathbb{L}}(x,y) = \operatorname{cosh}^{-1}\Big( x_0y_0 - \sum_{i=1}^m x_iy_i \Big)$.

\subsection{Network Architecture}
\label{sec:supp_image_generation_network_architecture}

The network architecture is given in Tab.~\ref{tab:network_architecture_image_generation}. 
The H-PROJ$_{\mathbb{L}_m \rightarrow \mathbb{B}_m}$ layer maps the Lorentz model into the Poincar\'e ball via the diffeomorphism 
given as
\begin{equation*}
\tau(x_0,x_1,\ldots,x_m) = \frac{(x_1,\ldots,x_m)}{x_0+1}. 
\end{equation*}

The H-PROJ$_{\mathbb{B}_m \rightarrow \mathbb{L}_m}$ layer maps the Poincar\'e ball into the Lorentz model via the diffeomorphism 
given as
\begin{equation*}
\tau^{-1}(x_1,\ldots,x_m) = \frac{(1 + \|x\|^2,2x_1,\ldots,2x_m)}{1 - \|x\|^2}. 
\end{equation*}

We briefly present the other layers below. Please refer to~\citet{BdeirFullyHnnCv24} for details. 

\paragraph{Lorentz FC layer}
\begin{equation*}
y = \operatorname{LFC}(x) = \begin{bmatrix} \sqrt{\| \rho(wx + b) \|^2 + 1} \\ \rho(wx + b) \end{bmatrix},
\end{equation*}
where $x,y$ are the input and output of the layer, respectively, 
$w \in \mathbb{R}^{m' \times (m+1)}$, and $b \in \mathbb{R}^{m'}$ and $\rho$ denote the bias and activation, respectively. 

\paragraph{Lorentz convolutional layer}
Given an image, the feature of each image pixel is mapped to the Lorentz model. 
Thus the image can be seen as an ordered set of $m$-dimensional hyperbolic feature vectors. 
The Lorentz convolution is then performed as
\begin{equation*}
y_{h,w} = \operatorname{LFC}(\operatorname{HCat}(\{ x_{h'+s\tilde{h},w'+s\tilde{w}} \}_{\tilde{h},\tilde{w}=1}^{\tilde{H},\tilde{W}})),
\end{equation*}
where $\{ x_{h'+s\tilde{h},w'+s\tilde{w}} \}_{\tilde{h},\tilde{w}=1}^{\tilde{H},\tilde{W}}$ are the features 
within the receptive field of the kernel, 
$\operatorname{HCat}(.)$ denotes the concatenation of hyperbolic vectors, 
$(h',w')$ denotes the starting position, and $s$ is the stride parameter. 

\paragraph{Lorentz transposed convolutional layer}
The transposed convolutional layer works by swapping the forward and backward passes of the convolutional layer. 
This is achieved in the Lorentz model through origin padding between the features.  

\paragraph{Lorentz batch normalization}
Given a batch $\mathcal{B}$ of $m$ features $x_i$, the traditional batch normalization algorithm can be described as
\begin{equation*}
\operatorname{BN}(x_i) = u \odot \frac{x_i - \operatorname{mean}(\mathcal{B})}{\sqrt{\operatorname{var}(\mathcal{B}) + \epsilon}} + v, 
\end{equation*}
where $\operatorname{mean}(\mathcal{B}) = \frac{1}{m}\sum_{i=1}^m x_i$, 
$\operatorname{var}(\mathcal{B}) = \frac{1}{m}\sum_{i=1}^m (x_i - \operatorname{mean}(\mathcal{B}))^2$, 
$u$ and $v$ are parameters to re-scale and re-center the features. 

For the Lorentz batch normalization layer, 
the Lorentzian centroid and the parallel transport operation are used for re-centering, 
and the Fr\'echet variance and straight geodesics at the origin's tangent space are used for re-scaling.

\paragraph{Lorentz ReLU}
\begin{equation*}
y = \begin{bmatrix} \sqrt{\| \operatorname{ReLU}([x_1,\ldots,x_m]) \|^2 + 1} \\ \operatorname{ReLU}([x_1,\ldots,x_m]) \end{bmatrix},
\end{equation*}
where $x=[x_0,\ldots,x_m]$ and $y$ are the input and output of the layer, respectively. 

\paragraph{Wrapped normal distribution}
The $\operatorname{SAMPLE}$ layer uses the method in~\citet{pmlrv97nagano19a} to generate random samples on $\mathbb{B}_m$. 
Given a normal distribution parameterized by a hyperbolic mean vector $h \in \mathbb{B}_m$ and a Euclidean 
variance matrix $\Sigma \in \mathbb{R}^{m \times m}$, the layer performs the following operations: 
\begin{enumerate}
\item Sample a Euclidean vector $\tilde{v}$ from the normal distribution $\mathcal{N}(0, \Sigma)$.
\item Compute $v = \frac{\tilde{v}}{2}$. 
\item Parallel transport $v$ from the tangent space of the origin $\mathbf{0}$ to the tangent space of the hyperbolic mean $h$ to 
obtain a tangent vector $u \in T_h \mathbb{B}_m$ as
\begin{equation*}
u = \mathcal{T}_{\mathbf{0} \rightarrow h}(v) = (1 - \|h\|^2)v. 
\end{equation*}
\item Map $u$ to $\mathbb{B}_m$ by applying the exponential map as
\begin{equation*}
z = \exp_h(u) = h \oplus_M \Big( \operatorname{tanh}\Big( \frac{\|u\|}{1 - \|h\|^2} \Big) \frac{u}{\|u\|} \Big), 
\end{equation*}
where $\oplus_M$ is the M\"{o}bius addition (see Appendix~\ref{sec:mobius_gyrovector_spaces}),  
and $z$ is the final sample in $\mathbb{B}_m$. 
\end{enumerate}

\subsection{Experimental Settings}
\label{sec:supp_image_generation_exp_settings}

\paragraph{Hyperparameters}
We adopt the hyperparameters from~\citet{BdeirFullyHnnCv24}. 
The curvature for the Lorentz model and the Poincar\'e ball is set to $1$. 
The learning rate and weight decay are set to $5e-4$ and $0$, respectively. 
The batch size and number of epochs are set to $100$. 
The KL loss weight is set to $0.024$. 

\paragraph{Optimization and evaluation} 
All models are implemented in Pytorch. We use the library Geoopt~\citep{geoopt2020kochurov} for Riemannian optimization. 
RiemannianAdam is used to train the networks. 
We use the reconstruction FID and generation FID to evaluate the networks. 
The reconstruction FID is computed by comparing test images with reconstructed validation images. 
A fixed random portion of $10$K images in the training set is used as the validation set~\citep{BdeirFullyHnnCv24}. 
The generation FID is computed by generating random images from the latent distribution and comparing them with the test set. 
Results are averaged over 5 runs for each model. 
We use a Quadro RTX 8000 GPU for all experiments. 

\subsection{Results}
\label{sec:supp_image_generation_results}

Results are shown in Tab.~\ref{tab:exp_hyp_networks_image_generation}. 
Our method achieves the best performances in terms of mean reconstruction FID and mean generation FID in all cases. 
We can also observe that the h-distance is significantly outperformed by the g-distance and b-distance in these experiments.  

\begin{table}[t]
\caption{\label{tab:exp_hyp_networks_image_generation} Reconstruction and generation FID of hyperbolic VAEs (lower is better).
}
\begin{center}
  \resizebox{0.95\linewidth}{!}{
  \def\arraystretch{1.3}
  \begin{tabular}{| l | c | c | c | c |}    
    \hline
    \multirow{2}{*}{Method} & \multicolumn{2}{c|}{CIFAR-10} & \multicolumn{2}{c|}{CIFAR-100} \\
    \cline{2-5}
    & Rec. FID & Gen. FID & Rec. FID & Gen. FID \\          
    \hline       
    Lorentz-Poincar\'e-H~\citep{fan2023horospherical} & 125.53$\pm$5.94 & 69.11$\pm$1.67 & 110.36$\pm$11.50 & 62.32$\pm$6.34 \\    
    Lorentz-Poincar\'e-G~\citep{NEURIPS2018_dbab2adc} & 39.68$\pm$1.45 & 49.91$\pm$2.06 & 42.82$\pm$2.48 & 60.24$\pm$4.01 \\
    \hline
    Lorentz-Poincar\'e-B (Ours) & {\bf 38.32}$\pm$2.11 & {\bf 48.45}$\pm$1.31 & {\bf 42.05}$\pm$2.58 & {\bf 59.76}$\pm$1.81 \\
    \hline	
  \end{tabular}
  } 
\end{center}
\end{table}

\section{Natural Language Inference}
\label{sec:supp_nli}

In this section, we compare our method for constructing the point-to-hyperplane distance in a Poincar\'e ball 
against those in~\citet{NEURIPS2018_dbab2adc,fan2023horospherical} by performing the same experiments in~\citet{NEURIPS2018_dbab2adc} 
for textual entailment and detection of noisy prefixes. For the first task, one has to predict 
whether a sentence can be inferred from another sentence. The second task consists of determining if a sentence is 
a noisy prefix of another sentence. 
Experiments are conducted on SNLI~\citep{bowman-etal-2015-large} and PREFIX datasets~\citep{NEURIPS2018_dbab2adc} for the first and second tasks, respectively.  
Our implementation\footnote{\url{https://github.com/sohata24/nli}.} is based on the open-source implementation\footnote{\url{https://github.com/dalab/hyperbolic_nn}.} 
of HypGRU~\citep{NEURIPS2018_dbab2adc} that uses the Poincar\'e MLR as a classification layer. 
The competing networks differ only in the computation of the point-to-hyperplane distance (see Tab.~\ref{tab:exp_distance_expressions}) 
in the Poincar\'e MLR. 

\subsection{Datasets}
\label{sec:supp_nli_datasets}

\paragraph{SNLI~\citep{bowman-etal-2015-large}}
It consists of $570$K training, $10$K validation and $10$K test sentence pairs. Similarly to~\citet{NEURIPS2018_dbab2adc},
The "contradiction" and "neutral" classes are merged into a single class of negative sentence pairs, 
while the "entailment" class gives the positive pairs.

\paragraph{PREFIX~\citep{NEURIPS2018_dbab2adc}}
PREFIX-10\%, PREFIX-30\%, and PREFIX-50\% are synthetic datasets, each of them consists of 
$500$K training, $10$K validation, and $10$K test pairs. Each dataset is built as follows. 
For each random first sentence of random length at most $20$ and one random prefix of it, 
a second positive sentence is generated by randomly replacing Z\% (Z is $10$, $30$, or $50$) of the words of the prefix, 
and a second negative sentence of same length is randomly generated.
Word vocabulary size is $100$. 

\begin{figure*}[t]
  \begin{center}
    \begin{tabular}{c}
      \includegraphics[width=0.7\linewidth, trim = 100 130 100 130, clip=true]{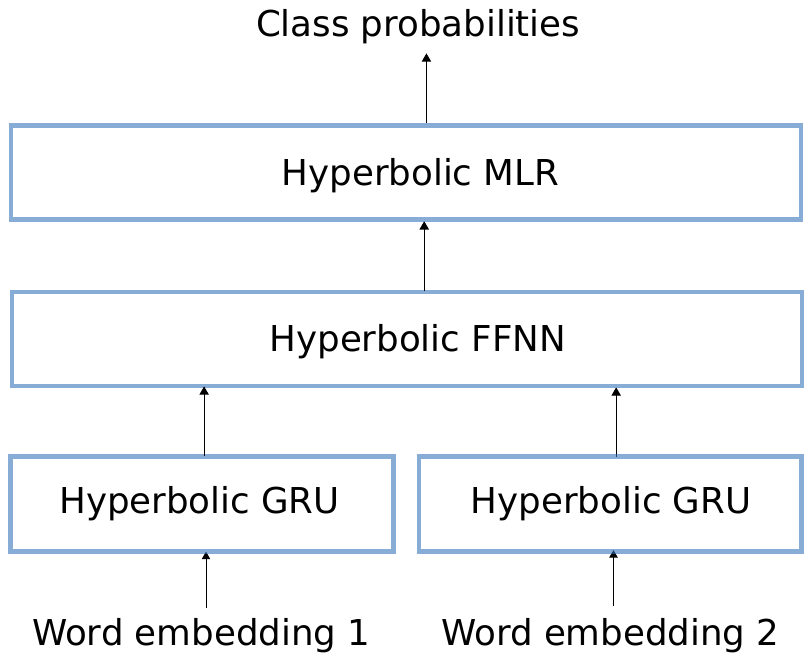} 
    \end{tabular}
  \end{center} 
  \caption{\label{fig:network_architecture_nli} The network architecture for natural language inference.}  
\end{figure*} 

\subsection{Experimental Settings}
\label{sec:supp_nli_exp_settings}

\paragraph{Network architecture}
We use the architecture of the fully hyperbolic GRU~\citep{NEURIPS2018_dbab2adc} illustrated in Fig.~\ref{fig:network_architecture_nli}.  
The network consists of a hyperbolic GRU, a hyperbolic feed forward neural network (FFNN), and the Poincar\'e MLR 
(see Appendix~\ref{sec:supp_image_classification_exp_settings}). 
The update equations of the hyperbolic GRU are given as
\begin{equation*}
r_t = \sigma\big( \log_{\mathbf{0}}( W^r \otimes_M h_{t-1} \oplus_M U^r \otimes_M x_t \oplus_M b^r ) \big),
\end{equation*}
\begin{equation*}
z_t = \sigma\big( \log_{\mathbf{0}}( W^z \otimes_M h_{t-1} \oplus_M U^z \otimes_M x_t \oplus_M b^z ) \big),
\end{equation*}
\begin{equation*}
\tilde{h}_t = \psi^{\otimes}( (W\operatorname{diag}(r_t)) \otimes_M h_{t-1} \oplus_M U \otimes_M x_t \oplus_M b),
\end{equation*}
\begin{equation*}
h_t = h_{t-1} \oplus_M \operatorname{diag}(z_t) \otimes_M (-h_{t-1} \oplus_M \tilde{h}_t). 
\end{equation*}
where $x_t \in \mathbb{B}_{m_1}$ is the input at frame $t$, 
$h_{t-1},h_t \in \mathbb{B}_{m_2}$ are the hidden states at frames $t-1$ and $t$, respectively, 
$W^r,W^z,W \in \mathbb{R}^{m_2 \times m_2}$, $b^r,b^z,b \in \mathbb{B}_{m_2}$, 
$U^r, U^z, U \in \mathbb{R}^{m_1 \times m_2}$ are model parameters, 
$\oplus_M$ is the M\"{o}bius addition (see Appendix~\ref{sec:mobius_gyrovector_spaces}), 
$\otimes_M$ is the M\"{o}bius matrix-vector multiplication (see Appendix~\ref{sec:mobius_gyrovector_spaces}), 
$\sigma,\psi$ are activation functions, 
$\log_{\mathbf{0}}(.)$ is the logarithmic map at $\mathbf{0}$ (see Appendix~\ref{sec:mobius_gyrovector_spaces}), 
and the function $\psi^{\otimes}(.)$ is defined as
\begin{equation*}
\psi^{\otimes}(x) = \exp_{\mathbf{0}}(\psi(\log_{\mathbf{0}}(x))),
\end{equation*}
where $x \in \mathbb{B}_m$ and $\exp_{\mathbf{0}}(.)$ is the exponential map at $\mathbf{0}$ (see Appendix~\ref{sec:mobius_gyrovector_spaces}). 

Let $x_1, x_2 \in \mathbb{B}_{m_2}$ be the outputs of the hyperbolic GRU corresponding to the first and second sentences, respectively. 
Then the output $z$ of the hyperbolic FFNN is computed as
\begin{equation*}
z = \tau( W_1 \otimes_M x_1 \oplus_M W_2 \otimes_M x_2 \oplus_M b_1 \oplus_M d_{\mathbb{B}}(x_1,x_2) \otimes_M b_2 ),
\end{equation*}
where $W_1, W_2 \in \mathbb{R}^{m_2 \times m_3}$, $b_1,b_2 \in \mathbb{B}_{m_3}$ are model parameters, 
$\otimes_M$ is the M\"{o}bius scalar multiplication (see Appendix~\ref{sec:mobius_gyrovector_spaces}), 
and $\tau$ is an activation function.   

\paragraph{Hyperparameters}
The word and hidden state embedding dimensions as well as the number of output channels of the hyperbolic FFNN are set to $5$. 
The number of epochs and batch size are set to $30$ and $64$, respectively. 
The learning rates for word embeddings and hyperbolic weights are set to $1e-1$ and $1e-2$, respectively. 
All activation functions in the hyperbolic GRU and hyperbolic FFNN are set to the identity function. 

\paragraph{Optimization and evaluation}
We use cross-entropy as the loss function. 
Euclidean and hyperbolic parameters are optimized using Adam (the learning rate is set to $1e-3$) 
and Riemannian stochastic gradient descent (RSGD)~\citep{BonnabelRSGD,pmlr-v80-ganea18a}, respectively. 
Results are averaged over $5$ runs for each model. 
We use a Quadro RTX 8000 GPU for all experiments. 

\subsection{Results}
\label{sec:supp_nli_results}

Results of the competing networks are shown in Tab.~\ref{tab:exp_hyp_networks_snli_prefix}. 
Our network outperforms HypGRU-H in terms of mean accuracy and standard deviation on all the datasets. 
Results also demonstrate that our proposed distance has the potential to improve existing HNNs on the considered task.  

\begin{table*}[t]
\caption{\label{tab:exp_hyp_networks_snli_prefix} Accuracies of the competing networks for natural language inference. 
}
\begin{center}
  \resizebox{0.95\linewidth}{!}{
  \def\arraystretch{1.2}
  \begin{tabular}{| l | c | c | c | c |}    
    \hline
    Method & SNLI & PREFIX-10\% & PREFIX-30\% & PREFIX-50\% \\          
    \hline
    HypGRU~\citep{NEURIPS2018_dbab2adc}           & 80.89$\pm$0.17 & 96.75$\pm$0.40 & 87.59$\pm$0.46 & {\bf 76.45}$\pm$0.61 \\ 
    HypGRU-H~\citep{fan2023horospherical} & 80.66$\pm$0.46 & 92.20$\pm$8.33 & 83.29$\pm$6.34 & 74.67$\pm$3.12 \\
    \hline
    HypGRU-B (Ours)                       & {\bf 81.01}$\pm$0.35 & {\bf 97.03}$\pm$0.11 & {\bf 87.69}$\pm$0.04 & 76.25$\pm$0.07 \\
    \hline
  \end{tabular}
  } 
\end{center}
\end{table*}

\section{Node Classification}
\label{sec:supp_exp_node_classification}

In this section, we compare our method for constructing the point-to-hyperplane distance in a Poincar\'e ball 
against those in~\citet{NEURIPS2018_dbab2adc,fan2023horospherical} by performing node classification experiments. 

\subsection{Datasets}

\paragraph{Disease~\citep{chami2019hyperbolic}}
It is the transductive variant of a dataset created by simulating the SIR disease spreading model where the label of a node is 
whether the node was infected or not, and node features indicate the susceptibility to the disease. 

\paragraph{Airport~\citep{chami2019hyperbolic}}
It is a flight network dataset from OpenFlights.org where nodes represent airports, edges represent the airline Routes, 
and node labels are the populations of the country where the airport belongs.   

\paragraph{Pubmed~\cite{namata:mlg12-wkshp}}
It is a standard benchmark describing citation networks where nodes represent scientific papers in the area of medicine, 
edges are citations between them, and node labels are academic (sub)areas. 

\paragraph{Cora~\cite{sen2008ccnd}}
It is a citation network where nodes represent scientific papers in the area of machine learning,  
edges are citations between them, and node labels are academic (sub)areas. 
Each publication in the dataset is described by a 0/1-valued word vector indicating the absence/presence of the 
corresponding word from the dictionary. 

The statistics of the four datasets are summarized in Tab.~\ref{tab:exp_node_classification_datasets}. 
Following~\citet{chami2019hyperbolic}, we use 30/10/60 percent splits for Disease dataset, 70/15/15 percent splits for Airport dataset, 
and standard splits~\citep{kipf2017semisupervised} with 20 train examples per class for Pubmed and Cora datasets.

\begin{table}[t]
\begin{center}
  \resizebox{0.55\linewidth}{!}{
  \def\arraystretch{1.3}
  \begin{tabular}{| l | c | c | c | c |}    
    \hline
    Dataset & \#Nodes & \#Edges & \#Classes & \#Features \\ 
    \hline       
    Disease  & 1044 & 1043 & 2 & 1000 \\
    Airport & 3188 & 18631 & 4 & 4 \\     
    Pubmed  & 19717 & 44338 & 3 & 500 \\      
    Cora     & 2708 & 5429 & 7 & 1433 \\   
    \hline	
  \end{tabular}
  } 
\end{center}
\caption{\label{tab:exp_node_classification_datasets} Description of the datasets for node classification experiments.}
\end{table}

\subsection{Experimental Settings}
\label{sec:supp_exp_node_classification_exp_settings}

\paragraph{Network architecture}
We use the HGCN architecture\footnote{\url{https://github.com/nguyenxuanson10/symspaces-nc}.} in~\citet{chami2019hyperbolic} which consists of a graph convolutional network (GCN) 
and a MLR as the final layer for classification. 
Both the GCN and the MLR layer are built on the Poincar\'e ball. 

\paragraph{Hyperparameters}
We follow closely the experimental settings in~\citet{chami2019hyperbolic}. The number of epochs is set to $5000$.
We use early stopping based on validation set performance with a patience of $100$ epochs. 
The learning rate is set to $1e-2$ for all experiments. 
The weight decays for Diease and Airport datasets are set to $0$. 
The weight decays for Pubmed and Cora datasets are set to $1e-4$ and $1e-3$, respectively. 
The number of dimensions is set to $16$. The number of layers in the GCN is set to $3$. 

\paragraph{Optimization and evaluation}
The network is implemented in Pytorch and is trained using cross-entropy loss and Adam optimizer. 
Results are averaged over $10$ random parameter initializations on the final test set. 

\subsection{Results}
\label{sec:supp_exp_node_classification_results}

Table~\ref{tab:exp_hyperbolic_node_classification} shows results of our experiments. 
We can observe that the b-distance gives the best performance in most cases, which is similar 
to our results on natural language inference experiments. It can also be observed that 
the performance of h-distance can nearly match those of g-distance and h-distance 
on less hyperbolic datasets (high hyperbolicity values), i.e., Pubmed and Cora datasets. 
However, the h-distance is significantly outperformed by the g-distance and b-distance 
on more hyperbolic datasets, i.e., Disease and Airport datasets. 
For those datasets, the b-distance achieves the best performances, demonstrating its effectiveness 
for datasets with stronger hierarchical structures. 
Furthermore, in all cases, the b-distance achieve the lowest standard deviation, 
suggesting that it can offer stable results in the considered application. 

\begin{table}[t]
\begin{center}
  \resizebox{0.75\linewidth}{!}{
  \def\arraystretch{1.3}
  \begin{tabular}{| l | c | c | c | c |}    
    \hline
    Dataset & Disease & Airport & Pubmed & Cora \\
    Hyperbolicity $\delta$ & $\delta=0$ & $\delta=1$ & $\delta=3.5$ & $\delta=11$ \\ 
    \hline   
    HGCN-H & 85.67 $\pm$ 2.58 & 69.82 $\pm$ 2.08 & 75.26 $\pm$ 1.82 & 77.09 $\pm$ 2.02 \\ 
    HGCN-G & 88.98 $\pm$ 1.96 & 84.78 $\pm$ 1.48 & {\bf 76.02 $\pm$ 1.09} & 77.47 $\pm$ 1.15 \\    
    \hline
    HGCN-B (Ours) & {\bf 89.05 $\pm$ 0.78} & {\bf 85.04 $\pm$ 0.97} & 75.89 $\pm$ 0.78 & {\bf 77.90 $\pm$ 1.00} \\        
    \hline	
  \end{tabular}
  } 
\end{center}
\caption{\label{tab:exp_hyperbolic_node_classification} Results of HGCN models for node classification. 
HGCN-H, HGCN-G, and HGCN-B are built on h-distance, g-distance, and b-distance, respectively. 
A lower hyperbolicity value $\delta$ means more hyperbolic. 
}
\end{table}

\section{Complexity Analysis}
\label{sec:complexity_analysis}

We analyze the complexities of the two network building blocks based on $G$-invariant metrics on SPD manifolds. Let $d_{in}$ and $d_{out}$ be 
the dimensions of input and output matrices of an FC layer, respectively. Let $l$ be the number of input matrices of the attention module.
\begin{itemize}
\item FC layer: It has memory complexity $O(d_{in}^2d_{out})$ and time complexity $O(d_{in}^3d_{out})$. 
\item Attention module: It has memory complexity $O(3d_{in}^2d_{out})$. The time complexity of the FC layers is $O(3ld_{in}^3d_{out})$. 
Both the ``Riemannian distance" and ``Midpoint operation" blocks have time complexity $O(l^2d_{out}^3)$. 
\end{itemize}

\section{Connection of Our Derived Point-to-hyperplane Distance with Existing Works}
\label{sec:supp_connection_random_features}

In the case where $\delta(t) = \exp(ta)K$, the distance given in Eq.~(\ref{eq:connection_to_composite_distance}) has a direct connection with 
the composite distance (see Section~\ref{sec:mathematical_background}). 
That is, the former is obtained by the action of functional $a \in \mathfrak{a}$ 
on the latter, which is a vector-valued distance. 
This is similar to how the Helgason-Fourier transform~\citep{helgason1994geometric} of a function on $X$ is formed. 
In particular, for any function $f$ on $X$, its Helgason-Fourier transform is defined as
\begin{equation*}
\tilde{f}(\lambda,\xi) = \int_X f(x)\exp( (-i\lambda + \varrho)A(x,\xi) ) dx,
\end{equation*}  
where $\lambda, \varrho \in \mathfrak{a}^{*}$, 
$\xi \in \partial X$, and $A(x,\xi) \in \mathfrak{a}$ denotes the composite distance from the origin $o$ to the horocycle passing 
through the point $x \in X$ with normal $\xi$. 
Here the exponent $(-i\lambda + \varrho)\langle x,\xi \rangle_{\mathbb{H}}$ is regarded as 
the action of functional $-i\lambda + \varrho \in \mathfrak{a}^{*}$ on the vector-valued distance 
$A(x,\xi)$~\citep{helgason1994geometric,Sonoda2022FCRidgele}. 

The distance given in Eq.~(\ref{eq:connection_to_composite_distance}) is also closely related to 
random features on hyperbolic spaces~\citep{YuICLR23}. Those features are generated from a map 
$\operatorname{HyLa}(.): \mathbb{B}_m \rightarrow \mathbb{R}$ given as
\begin{equation*}
\operatorname{HyLa}_{\lambda,b,\xi}(x) = \exp\Big( \frac{m-1}{2}A(x,\xi) \Big) \cos(\lambda A(x,\xi) + b),
\end{equation*}
where $x \in \mathbb{B}_m$, $\xi \in \partial \mathbb{B}_m$, and $\lambda,b \in \mathbb{R}$. 
By generating $m'$ random samples of tuple $(\lambda,b,\xi)$ from appropriate distributions, one 
obtains a feature map $\omega: \mathbb{B}_m \rightarrow \mathbb{R}^{m'}$ that approximates an isometry-invariant
kernel over hyperbolic space $\mathbb{B}_m$. In the higher-rank symmetric space setting, 
the term $\lambda A(x,\xi)$ will be replaced\footnote{Other adaptations are also needed but they are beyond the scope of our paper.} with the Euclidean inner product of a vector and a vector-valued distance, 
resulting in a similar formulation of the distance in Eq.~(\ref{eq:connection_to_composite_distance}).  

\section{Comparison of Our FC Layers against Existing Ones}
\label{sec:supp_fc_distinction}

In~\citet{HuangGool17,HuangAAAI18}, the authors introduce Bimap and FRMap layers and refer them as FC convolution-like layers. 
For both types of layers, each element of the output matrix is a linear combination of the elements of the input matrix,  
which is not the case in our FC layers. 
FRMap layers 
and our FC layers also differ in their outputs as the former do not produce points on the considered manifolds.    

In~\citet{ChakrabortyManifoldNet20}, 
the weighted Fr\'echet Mean (wFM) is adopted to develop Riemannian convolutional layers. 
These layers cannot be easily extended to build natural generalizations of Euclidean FC layers 
by simply treating FC layers as special cases of convolutional layers with full kernel size.
This is because the resulting FC layers will take as input a set of points on the considered manifold and therefore  
have no obvious connection with the linear transformation in Eq.~(\ref{eq:fc_euclidean}). 

FC layers for neural networks on hyperbolic spaces~\citep{wang2021laplacian} and symmetric spaces~\citep{Sonoda2022FCRidgele} 
include activation functions which are not used in our FC layers. 
Also, FC layers in~\citet{Sonoda2022FCRidgele,wang2021laplacian} do not output points on the considered spaces. 

In~\citet{NguyenICLR24,shimizu2021hyperbolic}, FC layers are not built upon Busemann functions. 
Although the method in~\citet{NguyenICLR24} is designed for matrix manifolds, some of which are not covered by our method 
(e.g., Grassmann manifolds),  
the former relies on differentiable forms of certain geometric quantities (e.g., the logarithmic map and parallel
transport) which are not required by the latter. 

Another line of work~\citep{cohen2018spherical,weiler2021coordinate} 
develops Riemannian neural networks which 
are functions $f: X \rightarrow \mathbb{R}^m$, where $m$ is the number of output channels. 
The context of these works is different from ours since we aim to build neural networks which are functions $f: X \rightarrow X$.

\section{Advantages and Limitations of $G$-invariant and PEM Metrics}
\label{sec:advantages_limitations}

Our motivation for using G-invariant metrics is that such metrics always exist on the considered
spaces (e.g., those induced from the Killing form~\citep{helgason1979differential}). As a result, the framework proposed in
Section~\ref{subsec:general_formulation_symspace} is valid for all those spaces. Since Busemann functions associated with $G$-invariant
metrics are determined (see the proof of Proposition~\ref{lem:busemann_function_symspaces}) by Busemann functions on the maximal
abelian $A$ of $G$ (recall that $G = KAN$) which has much lower dimension than $X$, a major advantage
of $G$-invariant metrics compared to PEM metrics is that the former allow FC layers to scale better
to high-dimensional input and output matrices. 
However, due to the same reason, Busemann functions associated with $G$-invariant metrics do not
fully capture the structure of $G$ which might lead to poor performance.

The use of PEM metrics is motivated by the fact that like Log-Euclidean metrics, they turn SPD
manifolds into flat spaces (i.e., their sectional curvature is null everywhere) which considerably simplifies 
computation and analysis (e.g., Riemannian computations become Euclidean computations in
the logarithmic domain~\citep{arsigny:inria-00070423}). PEM metrics are also more general than Log-Euclidean ones. On the
downside, those metrics do not yield full affine-invariance~\citep{arsigny:inria-00070423,Lin_2019}. This might break the geometric
stability principle that plays a crucial role in geometric deep learning architectures~\citep{BronsteinGDL}. 

\begin{figure}[t]
  \begin{center}
    \begin{tabular}{c}      
      \includegraphics[width=0.6\linewidth, trim = 100 100 70 130, clip=true]{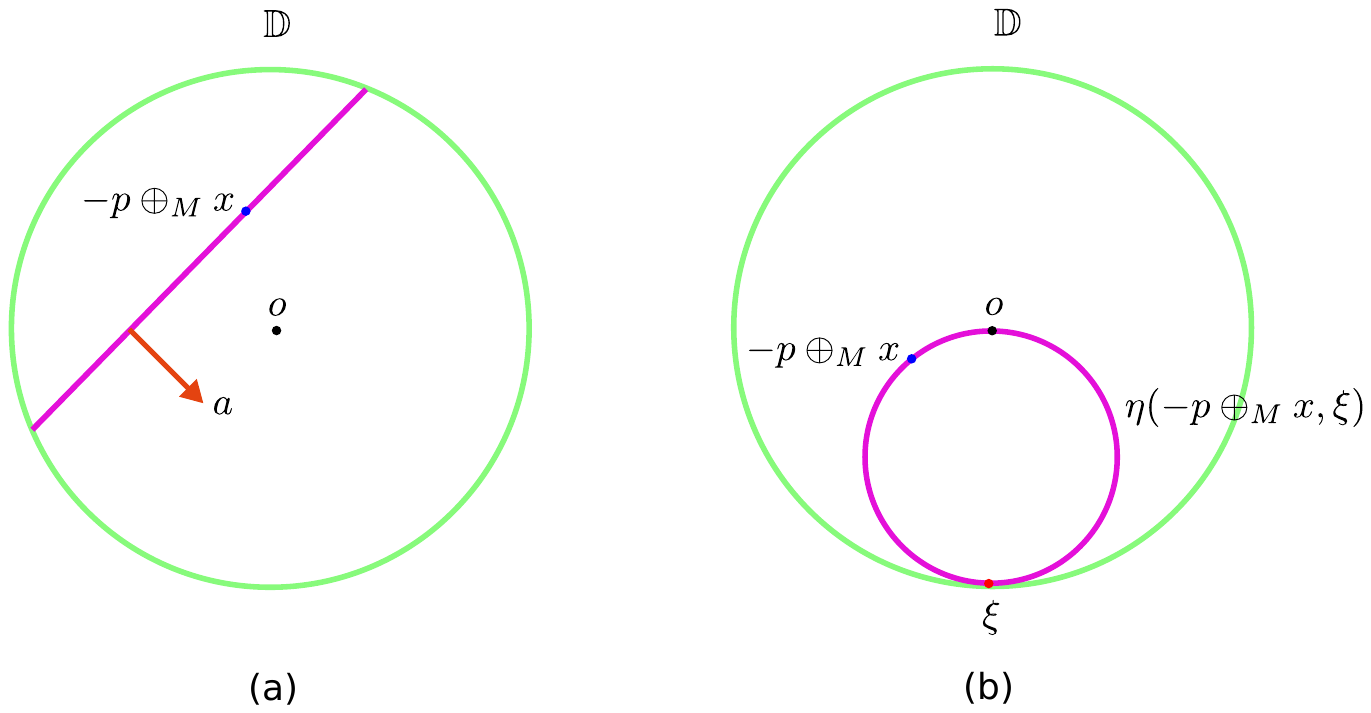} 
    \end{tabular}
  \end{center} 
  \caption{\label{fig:b_distance_vs_g_distance} Comparison of Poincar\'e hyperplanes and our hyperplanes in the Poincar\'e disk model.}.   
\end{figure}

\section{Comparison of the b-distance and g-distance}
\label{sec:supp_b_distance_vs_g_distance}

The difference between the b-distance and g-distance can be best explained 
in the Poincar\'e disk model $\mathbb{D}$ (see Appendix~\ref{sec:supp_hyperbolic_spaces_symspaces}).   
A Poincar\'e hyperplane~\citep{NEURIPS2018_dbab2adc} is defined as
\begin{equation*}
\mathcal{H}_{a,p} = \{ x \in \mathbb{D}: \langle \log_p(x),a \rangle_p = 0 \},
\end{equation*}
where $p \in \mathbb{D}$, $a \in T_p \mathbb{D} \setminus \{ \mathbf{0} \}$, 
and $\langle .,. \rangle_p$ is the Riemannian metric of the Poincar\'e disk model. 
It has been shown~\citep{NEURIPS2018_dbab2adc} that hyperplane $\mathcal{H}_{a,p}$ can also be described as
\begin{equation*}
\mathcal{H}_{a,p} = \{ x \in \mathbb{D}: \langle -p \oplus_M x,a \rangle = 0 \},
\end{equation*}
which is illustrated by the segment (in purple) in Fig.~\ref{fig:b_distance_vs_g_distance}(a). 

Since we use the M\"{o}bius addition $\oplus_M$ to define the binary operation $\oplus$, 
a hyperplane in our approach is characterized by
\begin{equation*}
\mathcal{H}_{\xi,p} = \{ x \in \mathbb{D}: B_{\xi}(-p \oplus_M x) = 0 \},
\end{equation*}
where $p \in \mathbb{D}$ and $\xi \in \partial \mathbb{D}$. One can interpret $B_{\xi}(-p \oplus_M x)$ as 
the signed distance between the origin $o$ and the horocycle $\eta(-p \oplus_M x,\xi)$ 
which is the unique horocycle~\citep{helgason1979differential} through $-p \oplus_M x$ with normal $\xi$. 
It implies that $o$ must lie on the horocycle $\eta(-p \oplus_M x,\xi)$. Hence, 
points $-p \oplus_M x$ must lie on the (unique) horocycle (in purple) through the origin $o$ with normal $\xi$ (see Fig.~\ref{fig:b_distance_vs_g_distance}(b)). 
It can be seen that the characterization of our hyperplanes and that of Poincar\'e hyperplanes are different, 
resulting in different formulations for the point-to-hyperplane distance. 

\section{Limitations of Our Approach}
\label{sec:supp_limitations}

In our work, the distance between a point and a hyperplane is derived for all higher-rank symmetric spaces of noncompact type, 
and the proposed FC layers are also designed for neural networks on those spaces. 
However, the attention module relies on the computation of wFM which does not have a closed-form solution 
in the general case. To address this issue, one can consider using the Karcher algorithm~\citep{KarcherRiemannianMean77} 
which has proven effective in the implementation of batch normalization layers in SPD neural networks~\citep{BrooksRieBatNorm19}. 
This algorithm 
can be computationally expensive in practice 
due to its iterative nature. To alleviate this challenge, the authors of~\citet{ChakrabortyManifoldNet20} introduced
an efficient wFM estimator which is worth investigating. 
One can also consider using the methods proposed in~\citet{LouwFMICML20}. 
In particular, the one that relies on an exponential map reparameterization~\citep{lezcano2019trivializations} 
can offer an effective solution to our problem. 

To obtain a closed form for the point-to-hyperplane distance, one has to derive a closed form for the Busemann function on the considered spaces and Riemannian metrics which is not always trivial.

Our FC layers have memory complexity $O(d_{in}^2d_{out})$ and time complexity $O(d_{in}^3d_{out})$. Even though our method significantly reduces the complexity of FC layers proposed in~\citet{NguyenICLR24}, it still does not scale well to high-dimensional input and output matrices of FC layers. Our attention block is built from these FC layers and thus suffers from the same issue. 

Finally, since our approach is developed for symmetric spaces of noncompact type, it cannot be applied to those of compact type, e.g., Grassmann manifolds that are also encountered in many machine learning applications. 

\section{Definitions and Basic Facts}
\label{sec:supp_definitions}

\subsection{Hyperbolic Spaces as Symmetric Spaces of Noncompact Type}
\label{sec:supp_hyperbolic_spaces_symspaces}

Here we describe the Poincar\'e disk model of the 2-dimensional hyperbolic space 
from a symmetric space perspective. 
Denote by $\mathbb{D} = \{ x \in \mathbb{C}: \|x\| < 1 \}$ the open unit disk in $\mathbb{C}$ equipped with
the Riemannian metric 
\begin{equation*}
\langle u,v \rangle_x = \frac{\langle u,v \rangle}{(1 - \| x \|^2)^2},
\end{equation*}
where $u,v \in T_x \mathbb{D}$ are tangent vectors at $x \in \mathbb{D}$. Let $G$ be the group defined as
\begin{equation*}
G = SU(1,1) := \left \{ \begin{bmatrix} a & b \\ \bar{b} & \bar{a} \end{bmatrix}: a,b \in \mathbb{C}, \|a\|^2 - \|b\|^2 = 1 \right \}. 
\end{equation*}

Let $SO_m$ be the group of $m \times m$ orthogonal matrices of determinant 1. Then $\mathbb{D}$ can be identified as
\begin{equation*}
\mathbb{D} \simeq SU(1,1)/SO_2.
\end{equation*}

The subgroups $K$, $A$, and $N$ in the Iwasawa decomposition $G = KAN$ and the centralizer $M$ of $A$ in $K$ are given by
\begin{equation*}
K = \left \{ \begin{bmatrix} e^{i\theta} & 0 \\ 0 & e^{-i\theta} \end{bmatrix}: \theta \in [0,2\pi) \right \},
\end{equation*}
\begin{equation*}
A = \left \{ \begin{bmatrix} \cosh t & \sinh t \\ \sinh t & \cosh t \end{bmatrix}: t \in \mathbb{R} \right \},
\end{equation*}
\begin{equation*}
N = \left \{ \begin{bmatrix} 1+is & -is \\ is & 1-is \end{bmatrix}: s \in \mathbb{R} \right \},
\end{equation*}
\begin{equation*}
M = \left \{ \begin{bmatrix} 1 & 0 \\ 0 & 1 \end{bmatrix}, \begin{bmatrix} -1 & 0 \\ 0 & -1 \end{bmatrix} \right \}.
\end{equation*}

The group $G$ acts on $\mathbb{D}$ by isometries via the M\"{o}bius action defined as 
\begin{equation*}
g[x] := \frac{ax + b}{\bar{b}x + \bar{a}}.
\end{equation*}

This map is conformal and maps circles and lines into circles and lines. 

\subsection{Pullback Euclidean Metrics}
\label{sec:supp_pem}

Under PEM, the Riemannian operations are given by:
\begin{equation*}
\exp_x(u) = \phi^{-1}(\phi(x) + D_x \phi(u)),
\end{equation*}
\begin{equation*}
\log_x(y) = D_{\phi(x)} \phi^{-1} (\phi(y) - \phi(x)),
\end{equation*}
\begin{equation*}
\mathcal{T}_{x \rightarrow y}(u) = D_{\phi(y)} \phi^{-1} \circ D_x \phi(u),
\end{equation*}
where $\exp_x(.)$, $\log_x(.)$, and $\mathcal{T}_{x \rightarrow y}(.)$ are the exponential map, logarithmic map, 
and parallel transport, respectively.

\subsection{SPD Manifolds as Symmetric Spaces of Noncompact Type}
\label{sec:supp_spd_manifolds_symspaces}

The SPD manifold $\operatorname{Sym}_m^+$ is a differentiable manifold of dimension $m(m+1)/2$. 
The tangent space $T_x \operatorname{Sym}_m^+$ at point $x \in \operatorname{Sym}_m^+$ of the manifold 
is isomorphic to $\operatorname{Sym}_m$. The Riemannian metric is given by
\begin{equation*}
\langle u,v \rangle_x = \operatorname{Tr}(x^{-1}ux^{-1}v),
\end{equation*}
where $u,v \in T_x \operatorname{Sym}_m^+$. This metric is $G$-invariant. 

Let $GL_m$ be the group of $m \times m$ invertible matrices,  
and let $O_m$ be the group of $m \times m$ orthogonal matrices. Then $\operatorname{Sym}_m^+$ can be identified as
\begin{equation*}
\operatorname{Sym}_m^+ \simeq GL_m/O_m.
\end{equation*}

Let $G = GL_m$. Then the subgroups $K$, $A$, and $N$ in the Iwasawa decomposition $G=KAN$ are given by
\begin{itemize}
\item $K = O_m$. 
\item $A$ is the subgroup of $m \times m$ diagonal matrices with positive diagonal entries.
\item $N$ is the subgroup of $m \times m$ upper-triangular matrices with diagonal entries $1$. 
\end{itemize}


Any $g \in G$ can be written as $g = kan$ for exactly one triple $(k,a,n) \in K \times A \times N$, 
and the map $K \times A \times N \rightarrow G$ sending $(k, a, n)$ to $kan$ is a diffeomorphism. 
The centralizer $M$ of $A$ in $K$ is $M := C_K(A) := \{ k \in K | ka = ak \text{ for all } a \in A \}$, 
which is the set of diagonal matrices with entries $\pm 1$. 
The (transitive) action of $G$ on $\operatorname{Sym}_m^+$ is defined as $g[x] := gxg^T$ for any $g \in G$ 
and $x \in \operatorname{Sym}_m^+$. 


 

\subsection{Symmetric Spaces of Noncompact Type}
\label{sec:supp_symspaces}

Symmetric spaces of compact type and of noncompact type are interchanged by Cartan duality. 
Each of those can be further categorized into two classes.

Symmetric spaces of compact type have non-negative sectional curvature. 
The two classes of symmetric spaces of compact type are:
\begin{itemize}
\item Homogeneous spaces of a compact Lie group defined by an involution (class 1). 
\item Compact Lie groups with bi-invariant metrics (class 2).
\end{itemize}

Symmetric spaces of noncompact type have non-positive sectional curvature and are diffeomorphic to Euclidean spaces. 
The two classes of symmetric spaces of noncompact type are:
\begin{itemize}
\item Homogeneous spaces of a noncompact, noncomplex Lie group, by a maximal compact subgroup (class 3, dual to class 1).
\item Homogeneous spaces of a complex Lie group by a real form (class 4, dual to class 2).
\end{itemize}
 
There is a correspondence between symmetric spaces of noncompact type and semisimple Lie groups with trivial centre and no compact factors: 
For any Lie group G with trivial centre and no compact factors, if we take a maximal compact subgroup $K$, 
then the quotient $X = G/K$ endowed with a G-invariant Riemannian metric forms a symmetric space of noncompact type. 

An important invariant of a symmetric space of noncompact type is its rank. 
The rank of a symmetric space $X$ of noncompact type is the maximum dimension of flats in $X$ 
(flats in $X$ are subspaces isometric to Euclidean spaces). A symmetric space of noncompact type can be
of rank one or of higher-rank. 
\begin{itemize}
\item Rank one symmetric spaces of noncompact type: The real, complex and quaternionic hyperbolic spaces 
and the hyperbolic plane over the Cayley numbers.
\item Higher-rank symmetric spaces of noncompact type: All symmetric spaces of noncompact type of rank greater than one. 
Typical examples are SPD manifolds. 
\end{itemize}

Many geometric properties of rank one symmetric spaces of noncompact type and those of higher-rank are distinct. 
Thus, it might be challenging to extend a method designed for hyperbolic spaces to the higher-rank symmetric space setting. 
For instance, in the Poincar\'e model, the point-to-hyperplane distance resulted from geodesic projections~\citep{ChamiICML21HoroPCA} 
can be computed in closed-form~\citep{NEURIPS2018_dbab2adc}. However, this is not the case for SPD manifolds (except those under specific Riemannian metrics, e.g. Log-Euclidean metrics)~\citep{NguyenGyroMatMans23}. 

\begin{figure}[t]
  \begin{center}
    \begin{tabular}{c}      
      \includegraphics[width=0.4\linewidth, trim = 410 150 100 150, clip=true]{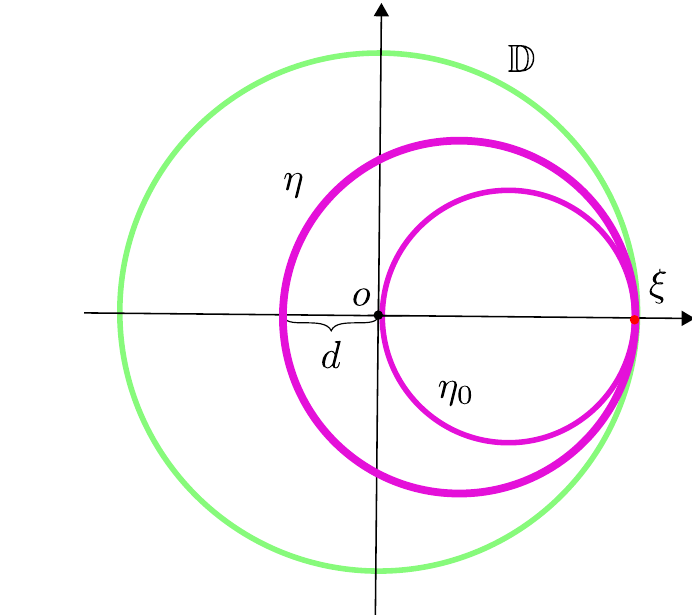} 
    \end{tabular}
  \end{center} 
  \caption{\label{fig:concepts_illustration} The boundary $\partial \mathbb{D}$ of the Poincar\'e disk model $\mathbb{D}$ is illustrated by the green circle. The boundary point $\xi$ is normal to both the horocycle $\eta$ and the basic horocycle $\eta_0$. The distance between the origin $o$ and the horocycle $\xi$ is $-d$.}   
\end{figure}

In the following, we provide several examples to illustrate some concepts reviewed in Section~\ref{sec:mathematical_background}. 

\paragraph{Geometric boundary}
Some examples of geometric boundaries of symmetric spaces of noncompact type are given below:

(1) For the Poincar\'e disk model of the $2$-dimensional hyperbolic space described in Section~\ref{sec:supp_hyperbolic_spaces_symspaces}, its geometric boundary $\partial \mathbb{D}$ is the unit circle $\partial \mathbb{D} = \{ x \in \mathbb{C}: \|x\|=1 \}$ (see Fig.~\ref{fig:concepts_illustration}). 

(2) Consider the Poincar\'e model $\mathbb{B}_m$ of $m$-dimensional hyperbolic geometry discussed in Section~\ref{sec:hyperbolic_space_spd_manifolds}. 
Two geodesic rays $\delta$ and $\delta'$ in $\mathbb{B}_m$ are asymptotic if and only if $\delta(t)$ and $\delta'(t)$ 
converge to the same point of $\mathbb{R}^m$ as $t \rightarrow \infty$. 
Thus the geometric boundary $\partial \mathbb{B}_m$ of $\mathbb{B}_m$ is naturally identified with the sphere of Euclidean radius $1$ 
centred at $\mathbf{0} \in \mathbb{R}^m$. 

(3) For $\operatorname{Sym}_m^+$, its geometric boundary $\partial \operatorname{Sym}_m^+$ consists of the left cosets
\begin{equation*}
\partial \operatorname{Sym}_m^+ := K/M := \{ \xi = kM | k \in K \}.
\end{equation*}

(4) If X is a complete $m$-dimensional Riemannian manifold of non-positive sectional curvature, 
then $\partial X$ is homeomorphic to $\mathbb{S}^{m-1}$, the $(m - 1)$-sphere. This is because for
a given base point $x \in X$, there exists a homeomorphism which associates to each unit vector $u$ 
tangent to $X$ at $x$ the class of the geodesic ray $\delta$ which issues from $x$ with velocity vector $u$. 




\paragraph{Horocycles} 
The horocycles of $\mathbb{D}$ and $\operatorname{Sym}_m^+$ can be described as follows.

(1) The geodesics in $\mathbb{D}$ are circular arcs perpendicular to the boundary $\partial \mathbb{D}$. All circular arcs perpendicular to the same point at $\partial \mathbb{D}$ can be seen as parallel lines.
Thus a natural notion of horocycle is that of circle tangent to $\partial \mathbb{D}$ (see Fig.~\ref{fig:concepts_illustration}). 

(2) Let $I_m$ be the $m \times m$ identity matrix. 
Then any horocycle $\eta$ can be written as $\eta = kaN[I_m]$ for some $k \in K$ and $a \in A$. 
In particular, it is the set of matrices having $a^2$ as diagonal matrix in the $UDU$ decomposition 
with respect to the $\mathbb{R}^m$-basis $\{ ke_i \}_{i=1,\ldots,m}$, 
where $\{ e_i \}_{i=1,\ldots,m}$ is the standard basis of $\mathbb{R}^m$~\citep{bartolucci2021unitarization}. 

\paragraph{Composite distances}
For $\operatorname{Sym}_m^+$, the composite distance $A(x,\xi)$ from the origin $o$ to the horocycle passing through a point 
$x \in \operatorname{Sym}_m^+$ with normal $\xi$ is given~\citep{Sonoda2022FCRidgele} by
\begin{equation*}
A(x=gK,\xi=kM) = \frac{1}{2} \log \gamma(k^T[x]),
\end{equation*}
where the map $\gamma: \operatorname{Sym}_m^+ \rightarrow A$ is determined by $y = v \gamma(y) v^T$ 
with $y \in \operatorname{Sym}_m^+$, $v \in N$. 

\subsection{Angles}
\label{sec:supp_angles}

\begin{definition}\label{def:comparison_angles}
A comparison triangle in $E^2$ for a triple of points $(p, q, r)$ in $X$ is a triangle in the Euclidean plane with vertices 
$\bar{p},\bar{q},\bar{r}$ such that $d(p,q) = d(\bar{p},\bar{q}), d(q,r) = d(\bar{q},\bar{r})$, and $d(p,r) = d(\bar{p},\bar{r})$. 
Such a triangle is unique up to isometry, and shall be denoted $\olsi{\Delta}(p,q,r)$. 
The interior angle of $\olsi{\Delta}(p,q,r)$ at $\olsi{p}$ is called the comparison angle between $q$ and $r$ at $p$ and is denoted 
$\olsi{\angle}_p(q,r)$. The comparison angle is well-defined provided $q$ and $r$ are both distinct from $p$. 
\end{definition}

\begin{definition}\label{def:geodesic_angles}
Let $\delta:[0,a] \rightarrow X$ and $\delta':[0,a'] \rightarrow X$ be two geodesic paths with $\delta(0) = \delta'(0)$. 
Given $t \in (0, a]$ and $t' \in (0, a']$, we consider the comparison triangle $\olsi{\Delta}(\delta(0),\delta(t),\delta'(t'))$, 
and the comparison angle $\olsi{\angle}_{\delta(0)}(\delta(t),\delta'(t'))$. The (Alexandrov) angle or the upper angle between the geodesic
paths $\delta$ and $\delta'$ is the number $\angle_{\delta,\delta'} \in [0,\pi]$ defined by:
\begin{equation*}
\angle_{\delta,\delta'} := \lim \sup_{t,t' \rightarrow 0} \olsi{\angle}_{\delta(0)}(\delta(t),\delta'(t')) = \lim_{\varepsilon \rightarrow 0} \sup_{0 < t,t' < \varepsilon} \olsi{\angle}_{\delta(0)}(\delta(t),\delta'(t')).
\end{equation*}
\end{definition}

\subsection{Busemann Functions}
\label{sec:supp_busemann_functions}

\paragraph{Euclidean spaces~\citep{bridson2011metric}}
Let $\delta(t) = tu$ be a ray in $\mathbb{R}^m$, where $u$ is a unit vector. 
Then the Busemann function $B_{\xi = \delta(\infty)}(.)$ is given by
\begin{equation*}
B_{\xi}(x) = -\langle x,u \rangle.
\end{equation*}


\paragraph{SPD manifolds under Log-Euclidean framework~\citep{BonetICML23SlicedWasserstein}}
Let $\delta(t) = \exp(ta)$ be a geodesic line in $\operatorname{Sym}_m^+$, 
where $a \in \operatorname{Sym}_m$ and $\| a \| = 1$. Then the Busemann function $B_{\xi = \delta(\infty)}(.)$ is given by
\begin{equation*}
B_{\xi}(x) = -\operatorname{Tr}(a \log(x)).
\end{equation*}

\subsection{Operations}
\label{sec:supp_operations}

\subsubsection{Poincar\'e Model and M\"{o}bius Gyrovector Spaces}
\label{sec:mobius_gyrovector_spaces}

In the Poincar\'e model $\mathbb{B}_m$ of $m$-dimensional hyperbolic geometry, 
the logarithmic map $\log_{\mathbf{0}}(.)$ and exponential map $\exp_{\mathbf{0}}(.)$ are given as
\begin{equation*}
\log_{\mathbf{0}}(x) = \operatorname{tanh}^{-1}(\|x\|) \frac{x}{\|x\|}, \hspace{3mm} \exp_{\mathbf{0}}(v) = \operatorname{tanh}(\|v\|) \frac{v}{\|v\|}, 
\end{equation*}
where $x\in \mathbb{B}_m \setminus \{ \mathbf{0} \}$ and $v \in T_{\mathbf{0}} \mathbb{B}_m \setminus \{ \mathbf{0} \}$.  

The M\"{o}bius addition $\oplus_M$ is defined as
\begin{equation*}
x \oplus_M y = \frac{ (1 + 2\langle x, y \rangle + \|y\|^2)x + (1 - \|x\|^2)y }{ 1 + 2\langle x, y \rangle + \|x\|^2 \|y\|^2 },
\end{equation*}
where $x, y \in \mathbb{B}_m$. The M\"{o}bius subtraction $\ominus_M$ is then defined as
\begin{equation*}
\ominus_M x = -x.
\end{equation*}

The M\"{o}bius scalar multiplication $\otimes_M$ is defined as
\begin{equation*}
r \otimes_M x = \operatorname{tanh}(r \operatorname{tanh}^{-1}(\| x \|)) \frac{x}{\| x \|},
\end{equation*}
where $r \in \mathbb{R}$ and $x \in \mathbb{B}_m \setminus \{ \mathbf{0} \}$.

The M\"{o}bius matrix-vector multiplication $\otimes_M$ is defined as
\begin{equation*}
M \otimes_M x = \operatorname{tanh}\bigg( \frac{\| xM \|}{\| x \|} \operatorname{tanh}^{-1}(\| x \|) \bigg) \frac{xM}{\| xM \|},
\end{equation*}
where $x \in \mathbb{B}_m$, $M \in \mathbb{R}^{m \times m'}$, and $M \otimes_M x = \mathbf{0}$ if $xM = \mathbf{0}$. 
Note that we use the same notation $\otimes_M$ for the M\"{o}bius scalar multiplication and 
M\"{o}bius matrix-vector multiplication as in~\citet{NEURIPS2018_dbab2adc}. 

\subsubsection{LE Gyrovector Spaces}
\label{sec:le_gyrovector_spaces}

For $x,y \in \operatorname{Sym}_m^{+}$, the binary operation $\oplus_{le}$ and inverse operation $\ominus_{le}$ 
are given as~\citep{NguyenECCV22,NguyenNeurIPS22}
\begin{equation*}
x \oplus_{le} y = \exp(\log(x) + \log(y)), \hspace{3mm} \ominus_{le} x = x^{-1},         
\end{equation*}
where $\exp(.)$ denotes the matrix exponential\footnote{As for function $\log(.)$, the meaning of function $\exp(.)$ should be clear from the context.}.

The SPD inner product is defined as
\begin{equation*}
\langle x,y \rangle^{le} = \langle \log(x), \log(y) \rangle.
\end{equation*} 


\section{Mathematical Proofs}
\label{sec:supp_proofs}

\subsection{Proof of Proposition~\ref{prop:distances_to_hyperplanes_pullback_metrics}}
\label{sec:supp_distances_to_hyperplanes_pullback_metrics}

\begin{proof}

We first recast a result from~\citet{chen2024riemannian} (Lemma 3.5) in form of the following proposition.

\begin{proposition}\label{prop:pull_metrics}
Let $\phi: \operatorname{Sym}_m^+ \rightarrow \operatorname{Sym}_m$ be a diffeomorphism,  
$p \in \operatorname{Sym}_m^+$, $a' \in T_p \operatorname{Sym}_m^+ \setminus \{ \mathbf{0} \}$,  
and let $\mathcal{H}^{pb}_{a',p}$ be the hyperplane defined as
\begin{equation*}
\mathcal{H}^{pb}_{a',p} = \{ x \in \operatorname{Sym}_m^+: \langle \operatorname{Log}_p(x),a' \rangle_p^{\phi} = 0 \}, 
\end{equation*}
where $\langle .,. \rangle_p^{\phi}$ is the PEM at point $p$ as given in the definition of SPD manifolds. 
Then the distance $d^{pb}(x,\mathcal{H}^{pb}_{a',p})$ between a point $x \in \operatorname{Sym}_m^+$ and 
hyperplane $\mathcal{H}^{pb}_{a',p}$ is given by
\begin{equation*}
d^{pb}(x,\mathcal{H}^{pb}_{a',p}) = \frac{| \langle \phi(x) - \phi(p), D_p \phi(a') \rangle |}{\| a' \|_p^{\phi}}, 
\end{equation*}
where $\|.\|_p^{\phi}$ is the norm induced by the Riemannian inner product $\langle .,. \rangle_p^{\phi}$. 
\end{proposition}

By the triangle inequality,
\begin{equation*}
d(\delta(0),\delta(t)) - d(x,\delta(0)) \le d(x,\delta(t)) \le d(\delta(0),\delta(t)) + d(x,\delta(0)),
\end{equation*}
which gives
\begin{equation*}
t - d(x,\delta(0)) \le d(x,\delta(t)) \le t + d(x,\delta(0)).
\end{equation*}

Thus
\begin{equation*}
1 - \frac{d(x,\delta(0))}{2t} \le \frac{d(x,\delta(t))+t}{2t} \le 1 + \frac{d(x,\delta(0))}{2t},
\end{equation*}
which leads to $\lim_{t \rightarrow \infty} \frac{d(x,\delta(t))+t}{2t} = 1$. 
Therefore
\begin{align*}
\begin{split}
B_{\xi=\delta(\infty)}(x) &= \lim_{t \rightarrow \infty} d(x,\delta(t)) - t \\ &= \lim_{t \rightarrow \infty} \big( d(x,\delta(t)) - t \big) \frac{d(x,\delta(t))+t}{2t} \\ &= \lim_{t \rightarrow \infty} \frac{1}{2t} \big( d(x,\delta(t))^2 - t^2 \big) \\ &=  \lim_{t \rightarrow \infty} \frac{1}{2t} \big( \| \phi(x) - \phi(\delta(t)) \|^2 - t^2 \big) \\ &= \lim_{t \rightarrow \infty} \frac{1}{2t} \big( \| \phi(x) \|^2 + \| \phi(\delta(t)) \|^2 - 2\langle \phi(\delta(t)),\phi(x) \rangle - t^2 \big) \\ &= \lim_{t \rightarrow \infty} \frac{1}{2t} \big( \| \phi(x) \|^2 + \| ta \|^2 - 2\langle ta,\phi(x) \rangle - t^2 \big) \\ &= \lim_{t \rightarrow \infty} \frac{1}{2t} \big( \| \phi(x) \|^2 - 2t \langle a,\phi(x) \rangle \big) \\ &= \lim_{t \rightarrow \infty} \frac{1}{2t} \big( \| \phi(x) \|^2 \big) - \langle a,\phi(x) \rangle \\ &= -\langle a,\phi(x) \rangle.  
\end{split} 
\end{align*}

By the definitions of the binary operation $\oplus$ and inverse operation $\ominus$, we have
\begin{equation*}
\ominus p \oplus x = \phi^{-1}(\phi(x) - \phi(p)).
\end{equation*}

Hence
\begin{equation*}
\| \ominus p \oplus x \|_{\mathbb{S}} = \| \phi(\ominus p \oplus x) \| = \| \phi(x) - \phi(p) \|.
\end{equation*}

We then get
\begin{align}\label{eq:equivalent_pull_metrics}
\begin{split}
\bar{d}(x,\mathcal{H}_{\xi,p}) &= d(x,p).\frac{B_{\xi}(\ominus p \oplus x)}{\| \ominus p \oplus x \|_{\mathbb{S}}} \\ &= -d(x,p).\frac{\langle a,\phi(\ominus p \oplus x) \rangle}{\| \phi(x) - \phi(p) \|} \\ &= -\langle a,\phi(\phi^{-1}(\phi(x) - \phi(p))) \rangle \\ &= -\langle a,\phi(x) - \phi(p) \rangle.
\end{split}
\end{align}

From Proposition~\ref{prop:pull_metrics},
\begin{equation*}
d^{pb}(x,\mathcal{H}^{pb}_{a',p}) = \frac{| \langle \phi(x) - \phi(p), D_p \phi(a') \rangle |}{\| a' \|_p^{\phi}} = \bigg| \langle \phi(x) - \phi(p), \frac{D_p \phi(a')}{\| a' \|_p^{\phi}} \rangle \bigg|,
\end{equation*}

By the property of pullback metrics,
\begin{equation*}
\langle a_1,a_2  \rangle_p^{\phi} = \langle D_p \phi(a_1),D_p \phi(a_2) \rangle,
\end{equation*}
where $a_1,a_2 \in T_p \operatorname{Sym}_m^+$. 
We deduce that
\begin{equation*}
\bigg\| \frac{D_p \phi(a')}{\| a' \|_p^{\phi}} \bigg\|= 1.
\end{equation*}

It can be seen that the unsigned distance $| \bar{d}(x,\mathcal{H}_{\xi,p}) |$ has precisely the same form as $d^{pb}(x,\mathcal{H}^{pb}_{a',p})$. 

\end{proof}

\subsection{Proof of Proposition~\ref{lem:main_inner_product_prop}}
\label{sec:supp_main_inner_product_prop}

\begin{proof}

To prove (i), we need a result from~\citet{kassel2009proper}.

\begin{lemma}\label{lem:complete_invariant_cartan_projection}
Let $\rho: X \rightarrow \olsi{\mathfrak{a}^+}$ be the map sending $x = g[o] \in X$ to $\mu(g)$, where $g \in G$. For all $x,x' \in X$,
\begin{equation*}
\| \rho(x) - \rho(x') \| \le d(x,x').
\end{equation*}

Moreover, if $x,x' \in \exp(\olsi{\mathfrak{a}^+})[o]$, then $d(x,x') = \| \rho(x) - \rho(x') \|$. 
\end{lemma}

Let $x = gK,y = hK$ where $g,h \in G$. Since the distance $d(.,.)$ is $G$-invariant, we have
\begin{align*}
\begin{split}
d(x,y) &= d(g^{-1}[x], g^{-1}[y]) \\ &= d(o, g^{-1}h[o]). 
\end{split}
\end{align*}

Let $g^{-1}h = kak'$ where $a \in \exp(\olsi{\mathfrak{a}^+})$, $k,k' \in K$. Then
\begin{align*}
\begin{split}
d(o, g^{-1}h[o]) &= d(k^{-1}[o], k^{-1}kak'[o]) \\ &= d(o, a[o])
\end{split}
\end{align*}

By Lemma~\ref{lem:complete_invariant_cartan_projection},
\begin{align*}
\begin{split}
d(o, a[o]) &= \| \rho(o) - \rho(a[o]) \| \\ &= \| \mu(a) \| \\ &= \| \mu(g^{-1}h) \|.
\end{split}
\end{align*}

Note that
\begin{align*}
\begin{split}
\| \ominus x \oplus y  \|_{\mathbb{S}} &= \| g^{-1}hK  \|_{\mathbb{S}} \\ &= \sqrt{\langle g^{-1}hK,g^{-1}hK \rangle_{\mathbb{S}}} \\ &= \sqrt{\langle \mu(g^{-1}h),\mu(g^{-1}h) \rangle} \\ &= \| \mu(g^{-1}h) \|.
\end{split}
\end{align*}

Therefore
\begin{equation*}
\| \ominus x \oplus y  \|_{\mathbb{S}} = d(x,y).
\end{equation*}

To prove (ii), note that for any $k \in K$, we have $k[x] = kgK = kk_1a_1n_1K = k_2a_1n_1K$ 
where $g = k_1a_1n_1$, $k_2=kk_1$, $k_1 \in K$, $a_1 \in A$, $n_1 \in N$. Thus $\mu(kg) = \mu(g)$. Similarly, 
we deduce that $\mu(kh) = \mu(h)$. Therefore
\begin{align*}
\begin{split}
\langle k[x],k[y] \rangle_{\mathbb{S}} &= \langle \mu(kg),\mu(kh) \rangle \\ &= \langle \mu(g),\mu(h) \rangle \\ &= \langle x,y \rangle_{\mathbb{S}}.
\end{split}
\end{align*}

\end{proof}

\subsection{Proof of Proposition~\ref{lem:busemann_function_symspaces}}
\label{sec:supp_busemann_function_symspaces}

\begin{proof}

We first recast a result from~\citet{bridson2011metric} (Lemma~10.26) in form of the following lemma.

\begin{lemma}\label{lem:n_leave_busemann_function_invariant}
Let $X$ be a symmetric space of noncompact type and let $G$ be a group acting by isometries on $X$. 
Suppose that $h \in G$ leaves invariant a geodesic line $\delta : \mathbb{R} \rightarrow X$ and that
$h[\delta(t)] = \delta(t + c)$ where $c > 0$. Let $x_0 = \delta(0)$ and let $N \subset G$ be the set of elements
$g \in G$ such that $h^{-j}gh^j[x_0] \rightarrow x_0$ as $j \rightarrow \infty$. Then 
$N$ fixes $\delta(\infty) \in \partial X$ and 
leaves invariant the Busemann function associated to $\delta$.
\end{lemma}

Let $\delta(t) = k\exp(ta)K$, $h = \exp(a) \in G$, where $a \in \mathfrak{a}$, $\|a\|=1$, $k \in K$. 
Setting $\delta'(t) = k^{-1}\delta(t)$. Then 
\begin{align*}
\begin{split}
h[\delta'(t)] &= \exp(a+ta)K \\ &= k^{-1}\delta(t+1) \\ &= \delta'(t+1).
\end{split}
\end{align*}

Since the distance $d(.)$ is G-invariant, for any $x \in X$, we have
\begin{equation*}
d(x,\delta(t)) = d(k^{-1}[x],\delta'(t)). 
\end{equation*}

Let $g \in G$ be such that $k^{-1}[x] = gK$, and let $g = n_1 \exp{A(g)} k_1$ where $n_1 \in N$, $k_1 \in K$\footnote{We use the same notation $A$ for the composite distance from the origin $o$ to a horocycle as in~\citet{helgason1994geometric}.}. 
For any $n \in N$, 
$h^{-j}nh^j[o] \rightarrow o$ as $j \rightarrow \infty$.
By Lemma~\ref{lem:n_leave_busemann_function_invariant} ($c=1$), we deduce that $B_{\xi'=\delta'(\infty)}(k^{-1}[x]) = B_{\xi'=\delta'(\infty)}(n_1^{-1}k^{-1}[x])$. 
Hence
\begin{equation*}
d(k^{-1}[x],\delta'(t))^2 = d(n_1^{-1}k^{-1}[x],\delta'(t))^2.
\end{equation*}

We thus have the following chain of equations
\begin{align}\label{eq:busemann_important_equality}
\begin{split}
d(k^{-1}[x],\delta'(t))^2 &= d(n_1 \exp{A(g)}K,\delta'(t))^2 \\ &= d(\exp{A(g)}K,\delta'(t))^2 \\ &= \langle A(g)-ta,A(g)-ta \rangle \\ &= \langle A(g),A(g) \rangle - 2t \langle a,A(g) \rangle + t^2. 
\end{split}
\end{align}

By the triangle inequality,
\begin{equation*}
d(\delta'(0),\delta'(t)) - d(k^{-1}[x],\delta'(0)) \le d(k^{-1}[x],\delta'(t)) \le d(\delta'(0),\delta'(t)) + d(k^{-1}[x],\delta'(0)),
\end{equation*}
which gives
\begin{equation*}
t - d(k^{-1}[x],\delta'(0)) \le d(k^{-1}[x],\delta'(t)) \le t + d(k^{-1}[x],\delta'(0)).
\end{equation*}

Thus
\begin{equation*}
1 - \frac{d(k^{-1}[x],\delta'(0))}{2t} \le \frac{d(k^{-1}[x],\delta'(t))+t}{2t} \le 1 + \frac{d(k^{-1}[x],\delta'(0))}{2t},
\end{equation*}
which results in $\lim_{t \rightarrow \infty} \frac{d(k^{-1}[x],\delta'(t))+t}{2t} = 1$. Therefore
\begin{align*}
\begin{split}
B_{\xi=\delta(\infty)}(x) &= \lim_{t \rightarrow \infty} d(x,\delta(t)) - t \\ &= \lim_{t \rightarrow \infty} d(k^{-1}[x],\delta'(t)) - t \\ &= \lim_{t \rightarrow \infty} \big( d(k^{-1}[x],\delta'(t)) - t \big) \frac{d(k^{-1}[x],\delta'(t))+t}{2t} \\ &= \lim_{t \rightarrow \infty} \frac{1}{2t} \big( d(k^{-1}[x],\delta'(t))^2 - t^2 \big). 
\end{split}
\end{align*}

Using Eq.~(\ref{eq:busemann_important_equality}), we get
\begin{align*}
\begin{split}
B_{\xi}(x) &= \lim_{t \rightarrow \infty} \frac{1}{2t} \big( \langle A(g),A(g) \rangle - 2t\langle a,A(g) \rangle \big) \\ &= -\langle a,A(g) \rangle \\ &= \langle a,H(g^{-1}) \rangle,
\end{split}
\end{align*}
which concludes Proposition~\ref{lem:busemann_function_symspaces}.

\end{proof}

\subsection{Proof of Corollary~\ref{corollary:distance_to_symspace_hyperplanes}}
\label{sec:supp_distance_to_symspace_hyperplanes}

\begin{proof}

We have
\begin{align*}
\begin{split}
\bar{d}(x,\mathcal{H}_{\xi,p}) &= d(x,p).\frac{B_{\xi}(\ominus p \oplus x)}{\|\ominus p \oplus x\|_{\mathbb{S}}} \\ &= d(x,p).\frac{B_{\xi}(h^{-1}gK)}{\|\ominus p \oplus x\|_{\mathbb{S}}}. 
\end{split}
\end{align*}

Note that $k^{-1}[h^{-1}gK] = k^{-1}h^{-1}gK$. By Propositions~\ref{lem:main_inner_product_prop} and~\ref{lem:busemann_function_symspaces},
\begin{align*}
\begin{split}
d(x,p).\frac{B_{\xi}(h^{-1}gK)}{\|\ominus p \oplus x\|_{\mathbb{S}}} &= \langle a, H(g^{-1}hk) \rangle,
\end{split}
\end{align*}
which concludes Corollary~\ref{corollary:distance_to_symspace_hyperplanes}. 

\end{proof}

\subsection{Proof of Proposition~\ref{prop:orthogonal_boundary_points}}
\label{sec:supp_orthogonal_boundary_points}

\begin{proof}

We first recast a result from~\citet{bridson2011metric} (Proposition 9.8) in form of the following proposition.

\begin{proposition}\label{prop:tits_metric_properties}
Let $X$ be a symmetric space of noncompact type with basepoint $x_0$. Let $\xi,\xi' \in \partial X$ and let $\delta,\delta'$ be 
geodesic rays with $\delta(0) = \delta'(0) = x_0$, $\delta(\infty) = \xi$ and $\delta'(\infty) = \xi'$. 
Then
\begin{equation*}
2\sin(\angle(\xi,\xi')/2) = \lim_{t \rightarrow \infty} \frac{1}{t} d(\delta(t),\delta'(t)). 
\end{equation*}
\end{proposition}

For any $t \in [0, \infty)$, we have that
\begin{align*}
\begin{split}
d(\delta(t),\delta'(t)) &= d(\exp(ta)K,\exp(ta')K) \\&= \| t(a - a') \| \\&= \sqrt 2t.
\end{split}
\end{align*}
By Proposition~\ref{prop:tits_metric_properties},
\begin{align*}
\begin{split}
2\sin(\angle(\xi,\xi')/2) &= \lim_{t \rightarrow \infty} \frac{1}{t} d(\delta(t),\delta'(t)) \\ &= \sqrt 2.
\end{split}
\end{align*}

We thus deduce that $\angle(\xi,\xi') = \frac{\pi}{2}$. 
\end{proof}

\subsection{Proof of Proposition~\ref{prop:fc_layers_symspaces}}
\label{sec:supp_fc_layers_symspaces}

\begin{proof}


Let $(e_j)_{j=1,\ldots,m}$ be the standard basis of $\mathbb{R}^m$, and let $\tilde{\xi}_j = \tilde{\delta}_j(\infty)$ 
where $\tilde{\delta}_j(t) = \exp(te_j)K,j=1,\ldots,m$ be geodesic rays. Then for $y=gK \in X$ and any $j \in \{ 1,\ldots,m \}$,
\begin{equation*}
v_j(x) = \bar{d}(y, \mathcal{H}_{\tilde{\xi}_j,K}) = \langle e_j, H(g^{-1}) \rangle = H(g^{-1})[j],
\end{equation*}
where $H(g^{-1})[j]$ denotes the $j$-th dimension of $H(g^{-1})$. Thus
\begin{equation*}
H(g^{-1}) = [v_1(x) \ldots v_m(x)]^T. 
\end{equation*} 
Note that $g = n \exp (-H(g^{-1})) k$ with $n \in N$ and $k \in K$. Therefore
\begin{equation*}
g = n \exp([-v_1(x) \ldots -v_m(x)]) k,
\end{equation*}
which leads to
\begin{equation*}
y = n \exp([-v_1(x) \ldots -v_m(x)]) K.
\end{equation*}

\end{proof}

\end{document}